\documentclass{bmvc2k}

\title{\datasetlong: Benchmarking Answer Grounding in Document Visual\\Question Answering} 

\addauthor{Luca De Grandis}{luca.degrandis@unimore.it}{1}
\addauthor{Silvia Cappelletti}{silvia.cappelletti@unimore.it}{1}
\addauthor{William Raccagni}{william.raccagni@unimore.it}{1,2}
\addauthor{Marcella Cornia}{marcella.cornia@unimore.it}{1}
\addauthor{Lorenzo Baraldi}{lorenzo.baraldi@unimore.it}{1}
\addauthor{Rita Cucchiara}{rita.cucchiara@unimore.it}{1}

\addinstitution{University of Modena and\\Reggio Emilia\\
Modena, Italy}
\addinstitution{University of Pisa\\
Pisa, Italy}

\runninghead{De Grandis et al.}{Benchmarking Answer Grounding in Document VQA}

\def\eg{\emph{e.g}\bmvaOneDot}

\def\etal{\emph{et al}\bmvaOneDot}

\usepackage{booktabs}
\usepackage{amsfonts}
\usepackage{nicefrac}
\usepackage{microtype}
\usepackage{xcolor}
\usepackage{graphicx}
\usepackage{amsmath}
\usepackage{amssymb}
\usepackage{array}
\usepackage{multirow}
\usepackage{color}
\usepackage{colortbl}
\usepackage{framed}
\usepackage{bm}
\usepackage{xspace}
\usepackage{enumitem}
\usepackage{xparse}
\usepackage{algorithm}
\usepackage{listings}
\usepackage{url}
\usepackage{mathtools}
\usepackage{pifont}
\usepackage{mathtools}
\usepackage{lipsum}
\usepackage{adjustbox}
\usepackage{wrapfig}

\usepackage{makecell}

\makeatletter
\newcommand{\manualtablecaption}[1]{
\normalsize
  \refstepcounter{table}
  \@makecaption{\tablename~\thetable}{#1}
}
\makeatother

\usepackage{placeins}
\usepackage{afterpage}

\usepackage{tcolorbox}
\definecolor{promptbg}{rgb}{0.95,0.95,0.95}
\definecolor{promptborder}{rgb}{0.85,0.85,0.85}

\definecolor{Gray}{gray}{0.2}
\definecolor{lightgray}{gray}{0.92}
\definecolor{TitleColor}{gray}{0.95}

\definecolor{blond}{rgb}{0.98, 0.94, 0.75}
\definecolor{LightCyan}{rgb}{0.88,0.95,1}
\definecolor{OurColor}{rgb}{0.855, 0.937, 0.957}

\definecolor{myhighlight}{HTML}{C7EDD6}

\def \ie {\emph{i.e.}}
\def \eg {\emph{e.g.}}
\def \etal {\emph{et al.}}

\newcommand{\tit}[1]{\smallbreak\noindent\textbf{#1.}}
\newcommand{\tinytit}[1]{\noindent\textbf{#1.}}

\newcommand{\ours}{MAPPET\xspace}

\newcommand{\datasetlong}{DocAttriBench\xspace}
\newcommand{\dataset}{DAB\xspace}

\newcommand{\task}{VAG\xspace}
\newcommand{\tasklong}{Visual Answer Grounding\xspace}

\newcommand{\vsa}{MAPPET}

\begin{document}
\sloppy

\maketitle

\begin{abstract}
Answer grounding in document visual question answering remains an open challenge: most benchmarks lack grounding annotations or provide limited-quality labels, while constructing grounded datasets still requires costly manual effort. We introduce \datasetlong (\dataset), a large-scale benchmark for fine-grained, element-level source attribution in Document VQA, grounding answers to specific layout elements such as text blocks, tables, and images. To build \dataset, we propose a Mask-based Perplexity-Derived Attribution method (\ours) that combines document layout and language modeling to identify the most informative element for each answer. \ours measures the increase in perplexity after masking candidate elements and attributes the answer to the element contributing most to model confidence. Applying \ours to multiple existing Document VQA datasets yields \dataset, with 237k documents and 296k question-answer pairs with element-level grounding. We benchmark grounding-capable multimodal LLMs on \dataset, evaluating answer accuracy, attribution accuracy, and overall answer quality. Results show that while larger models generally achieve higher answer accuracy, even the strongest models often fail to localize the supporting elements. \dataset provides a scalable benchmark for developing grounded, verifiable, and trustworthy Document VQA models. Dataset and code are available at \url{https://aimagelab.github.io/DocAttriBench/}.
\end{abstract}
\section{Introduction}
\label{sec:intro}

\begin{figure*}
    \centering
    \includegraphics[width=\linewidth]{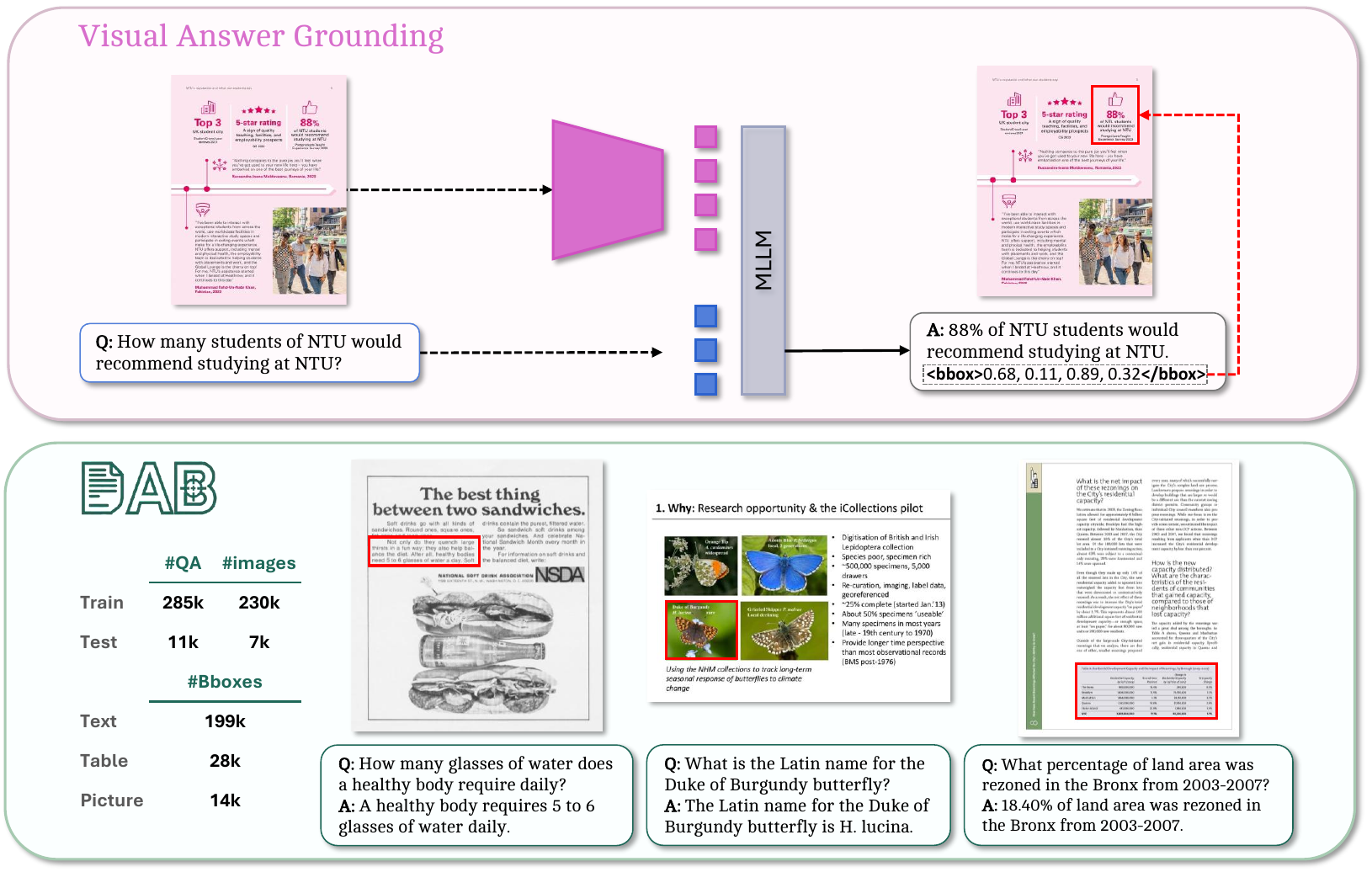}
    \vspace{-0.6cm}
    \caption{\textbf{(Top)} The \tasklong (\task) task, which our method advances. Specifically, a Multimodal Large Language Model (MLLM) answers an image query with text and a bounding box of the evidence region. \textbf{(Bottom)} Examples from our dataset, \datasetlong (\dataset) with dataset statistics: number of question-answer pairs (\#QA), unique images, and bounding boxes per type (\ie, text, table, and picture).}
    \label{fig:vag_dab}
    \vspace{-0.5cm}
\end{figure*}

\tasklong (\task) refers to the task of identifying the visual evidence that supports a given answer in a Visual Question Answering (VQA) setting~\cite{Antol_2015_ICCV, wu2017visual, Chen_2022_CVPR, Zhu_2016_CVPR}. Specifically, a model is required not only to produce a textual answer, but also to localize the region of the image that justifies it. Modern large-scale multimodal models~\cite{vlm_survey_zhang,caffagni2024r,Liu_2024_CVPR,bucciarelli2024personalizing,compagnoni2026reag}, built upon Transformer-based architectures~\cite{attention_is_all_you_need_nips_vaswani}, have achieved state-of-the-art performance, leading to the emergence of several families of foundation models designed to enable increasingly stronger multimodal reasoning.

In the context of document understanding, answer grounding was initially explored at the textual level by prompting Large Language Models (LLMs)~\cite{llm_overview_naveed,cappelletti2025improving} to generate verifiable citations supporting their answers, either through prompt engineering~\cite{gao-etal-2023-enabling} or supervised fine-tuning~\cite{ye-etal-2024-effective}. The growing capabilities of Multimodal Large Language Models (MLLMs)~\cite{caffagni2024r} have subsequently enabled a natural extension of this task to the visual domain, giving rise to \tasklong~\cite{ma2024mmlongbench, ma-etal-2025-visa, deng-etal-2025-longdocurl}, a more intuitive and rapidly verifiable approach. Despite this progress, visual answer grounding in documents remains an open challenge. Evidence from non-document domains suggests that, while MLLMs can achieve strong answer accuracy, their grounding capabilities remain limited~\cite{suris2023vipergpt}.

Recent works show that fine-tuning foundation MLLMs on Document VQA datasets annotated with grounding information significantly improves attribution performance~\cite{ma-etal-2025-visa} compared to non-specialized models~\cite{bai2025qwen25vltechnicalreport, Chen_2024_CVPR, chen2024far, chen2024expanding, zhu2025internvl3exploringadvancedtraining}. However, most existing benchmark datasets primarily focus on the VQA task itself, providing partial, coarse, inconsistently verified or noisy grounding annotations~\cite{ma-etal-2025-visa,tanaka2021visualmrc, deng-etal-2025-longdocurl, van2023document, giovannini2025boundingdocs}, no explicit link between answers and supporting visual evidence~\cite{liu2024visualwebbench, ma2024mmlongbench, tanaka2023slidevqa, wiedmann2025finevision}, or remaining limited in scale~\cite{ma2026citevqa}. High-quality grounding annotations are essential not only for assessing the alignment between model responses and the supporting content, but also for training more reliable models and mitigating hallucinations~\cite{mishra2024finegrained}. Constructing such datasets remains a non-trivial challenge. Manual annotation is costly and labor-intensive, requiring human annotators to identify and verify the visual evidence that supports each answer~\cite{tanaka2021visualmrc}. Automatic approaches can reduce human effort but may introduce errors or hallucinated grounding, where the linked visual content does not faithfully support the answer~\cite{min-etal-2023-factscore, gao-etal-2023-enabling, malaviya-etal-2024-expertqa}. Furthermore, many automated annotation approaches rely on proprietary or commercial APIs, which increases financial costs and can limit scalability for large-scale dataset construction~\cite{deng-etal-2025-longdocurl}.

Building on these observations, we develop two complementary components. First, we present \datasetlong (\dataset), a large-scale Document VQA dataset with high-confidence, element-level visual answer grounding annotations, designed to enable fine-grained evaluation of attribution performance and effective fine-tuning of MLLMs. Second, we use \textbf{Ma}sk-based \textbf{P}er\textbf{p}l\textbf{e}xity-derived A\textbf{t}tribution (\ours), a novel framework for automatic visual answer grounding, to construct \dataset. \ours performs two key functions: (i) localizing the spatial coordinates of the visual region that supports a given answer, and (ii) filtering low-quality or ambiguous samples by applying thresholds on the attribution score and score margin between candidate elements. 

\datasetlong is constructed by aggregating existing Document VQA datasets and processing them through our automatic annotation pipeline. It comprises 237,418 document images spanning diverse domains such as scientific articles, business reports, and presentation slides, along with 295,554 sets of image, question, answer, and bounding box, covering visual elements like text paragraphs, tables, and figures. For each combination of document, question, and answer, a document layout analysis model first extracts all the structural elements positions. \ours then iteratively masks layout elements and computes a perplexity-based score on the masked image. The attribution score is defined as the difference between the masked and unmasked perplexity scores, and the element with the highest score is selected as supporting evidence, provided it meets the pipeline filtering criteria. This process converts heterogeneous Document VQA sources into a unified element-level grounding benchmark. Figure~\ref{fig:vag_dab} depicts both the \task task and \datasetlong.

Extensive experiments validate \ours's ability to produce reliable annotations and the usefulness of DocAttriBench as a ground-centric benchmark. Human evaluation shows that \ours produces high-quality attributions, with accuracy exceeding 95\% for some benchmarks after automatic filtering, validating its ability to discard noisy and ambiguous examples. We further assess whether the selected evidence is sufficient on its own and necessary for maintaining answer log-likelihood. Through the evaluation of sixteen state-of-the-art MLLMs, we observe that zero-shot grounding is consistently weak, with the grounding score rarely surpassing 35\% for the largest models ($>$30B) and often falling below 10\% for the smaller ones ($\le$3B). Fine-tuned variants outperform their zero-shot counterparts by a large margin, even surpassing larger models in grounding accuracy, demonstrating that \dataset provides robust supervision to develop grounding-capable MLLMs.

\tit{Contributions} Our main contributions are summarized as follows:
\begin{itemize}[noitemsep, topsep=0pt]
    \item We introduce \dataset, a large-scale benchmark and training dataset comprising 237k document images and 296k grounded question-answer examples across multiple document types (scientific papers, reports, slides, and others), providing high-quality annotations for precise evaluation and fine-tuning of grounding-aware MLLMs.
    \item We propose \ours, a novel framework for automatic VAG that localizes answer-supporting regions and filters low-quality samples via confidence-based thresholding. At its core, \ours employs a Mask-based Perplexity-Derived Attribution mechanism that quantifies the contribution of each document layout element through perplexity variation induced by masking.
    \item Through extensive experiments on sixteen MLLMs, we demonstrate that supervision derived from \dataset consistently improves grounding  performance. Fine-tuned 7B-8B models trained on our data outperform their zero-shot counterparts by over 20 points and surpass larger  general-purpose models (32B-38B), confirming the effectiveness of our dataset and attribution pipeline for developing grounding-capable MLLMs.
\end{itemize}

\section{Related Work}
\label{sec:related}
\tinytit{Document Visual Understanding}
Document visual understanding aims to jointly reason over textual, visual, and layout cues in scanned or digital-born documents~\cite{subramani2020survey, layout_survey}. Recently, large-scale multimodal models~\cite{vlm_survey_zhang,caffagni2024r} have become the leading approach in document visual understanding, jointly modeling visual layouts, text, and spatial relations within a unified multimodal space. Powered by large-scale image-text pretraining and transformer architectures, they enable rich cross-modal reasoning.

Recent advances focus on OCR-free models~\cite{ye2023mplug, yu2025visrag}, which infer textual and structural semantics directly from images. These operate on high-resolution images, developing various strategies to manage the resulting computational load in order to improve document comprehension capabilities. End-to-end approaches such as Donut~\cite{donut_kim_2022}, PaLI-X~\cite{chen2023pali}, and Qwen2-VL~\cite{wang2024qwen2} aim to process full-resolution document images directly, yet typically rely on image downscaling to maintain computational tractability. In contrast, tile-based models like UReader~\cite{ye-etal-2023-ureader} and InternVL2~\cite{chen2024far} enhance efficiency by partitioning documents into smaller regions processed independently. Hybrid strategies, as adopted in LLaVA-1.5~\cite{Liu_2024_CVPR} and LLaVA-OneVision~\cite{li2024llava}, preserve global coverage by tiling full-resolution inputs while subsequently downsampling the aggregated visual representations.

\tit{Context Attribution}
Context attribution remains an open challenge, with recent efforts extending beyond textual grounding to visual contexts. On the textual side, Gao~\etal~\cite{gao-etal-2023-enabling} introduced ALCE, an automatic benchmark for evaluating LLMs ability to generate text with verifiable citations, providing reproducible metrics for fluency, correctness, and citation quality to mitigate hallucinations. Ye~\etal~\cite{ye-etal-2024-effective} proposed AGREE, a learning-based framework that fine-tunes LLMs for grounded and citation-accurate responses using automatically constructed data, coupled with test-time adaptation to iteratively retrieve additional evidence, thus enhancing factual reliability. ContextCite~\cite{cohen2024contextcite} attributes text through sampled context ablations and a sparse LASSO surrogate over response-probability changes. MAPPET instead directly scores masked spatial document elements and filters weak or ambiguous attributions. 
Extending attribution to the visual domain, ChartCitor~\cite{goswami2025chartcitor} applies context attribution to chart understanding by linking generated answers to specific visual evidence, using a multi-agent pipeline that maps answer claims to table cells and then grounds them in chart elements through localized visual citations. Ma~\etal~\cite{ma-etal-2025-visa} presented VISA, a vision-based RAG (V-RAG) framework that leverages MLLMs to jointly generate answers and visually localize supporting evidence within document screenshots. Built upon Qwen2-VL and fine-tuned on the constructed datasets, VISA achieves fine-grained visual grounding and establishes a new baseline for visual context attribution. More recently, CiteVQA~\cite{ma2026citevqa} studied evidence attribution in long-form Document VQA through an automated pipeline that constructs candidate evidence packages and uses element-level ablation to determine which components are crucial for answer recovery. In contrast, MAPPET uses a perplexity-based masking criterion to assign answers
to the most informative layout elements, enabling scalable grounding of existing Document
VQA datasets.

\tit{Document VQA Benchmarks}
Several benchmarks have been introduced for visual question answering on documents, but most suffer from limited scale, weak grounding, or low diversity. DocVQA~\cite{Mathew_2021_WACV} focuses on scanned UCSF documents but offers full annotations only for its 5k validation set and lacks bounding boxes. VisualMRC~\cite{tanaka2021visualmrc} provides 30k QA pairs over webpage screenshots with OCR and ROI annotations, but its construction relies on more than 500 human annotators for document selection, ROI labeling, and QA creation, making the process expensive, slow, and prone to inconsistencies. Moreover, the collected pages are generally short and structurally simple. LongDocURL~\cite{deng-etal-2025-longdocurl} extends to multi-page documents but has misaligned boxes and limited verification. It contains only 2.3k QA pairs and relies on proprietary models, including GPT-4o for document-type classification and QA generation, as well as commercial tools (\eg, DocMind) for PDF parsing and layout extraction.

DoclingMatix~\cite{nassar2025smoldocling} provides 2.4M synthetic instruction-response pairs without grounding annotations or page-level attributions. MMLongBench-Doc~\cite{ma2024mmlongbench} targets long-context multi-modal reasoning across 135 PDF documents with 1k expert-annotated questions involving text, tables, and figures; while it introduces cross-page and unanswerable questions, the dataset remains small and lacks spatial grounding.
DUDE~\cite{van2023document} is large-scale and covers diverse real-world documents and answer formats (extractive, abstractive, list-based, and non-answerable), but spatial grounding is provided only for extractive answers, all annotations are human-generated, and test-set evaluation relies on a submission server with gold answers withheld. Our analysis of its validation split reveals inconsistencies, including incorrect bounding boxes and misreferenced pages or images. BoundingDocs~\cite{giovannini2025boundingdocs} unifies multiple business-document datasets into QA pairs with word-level bounding boxes using Amazon Textract, but it supports only extractive answers and relies solely on text.

Structured resources such as the VISA family~\cite{ma-etal-2025-visa} provide large-scale VQA datasets with varying grounding quality. Wiki-VISA (90k) grounds human-annotated question-answer pairs on webpage elements, whereas Paper-VISA (100k) and FineWeb-VISA (60k) generate synthetic QA pairs by prompting VLMs with bounding box-overlaid screenshots. Paper-VISA includes unnatural questions referencing the bounding box itself and contains a non-negligible fraction of incorrect groundings, while FineWeb-VISA lacks human annotation, potentially reducing data quality. SlideVQA~\cite{tanaka2023slidevqa} (14k QAs) covers 14k QA pairs across 52k slide images, supporting single-hop, multi-hop, and numerical reasoning (25.5\% arithmetic), but does not include bounding boxes and provides limited reasoning supervision. VisualWebBench~\cite{liu2024visualwebbench} WebQA split contains only 314 webpage QA samples without bounding box annotations.
CiteVQA~\cite{ma2026citevqa} introduces a long-form Document VQA benchmark with element-level
visual citations, covering 1,897 questions over 711 multi-page PDFs and evaluating answer
correctness jointly with evidence faithfulness. While it targets faithful attribution evaluation
in long-document settings, DocAttriBench provides a substantially larger resource, supporting
both standardized benchmarking and fine-tuning of grounding-capable MLLMs. Overall, existing Document VQA datasets either remain small, lack spatial grounding, rely heavily on human annotation or proprietary tools. These limitations hinder large-scale training and robust evaluation of models for visually grounded QA.

\datasetlong addresses these gaps with large-scale, fine-grained annotations over diverse document types. It provides QA pairs linked to precise bounding boxes and region types, ensuring strong visual-textual alignment. This enables robust benchmarking and effective training for visually grounded VQA on real-world documents.

\section{\datasetlong (\dataset)}
\label{sec:dataset}

\subsection{Dataset Overview}\label{sec:dataset_overview}
\datasetlong is a large-scale benchmark dataset designed to evaluate MLLMs on \task in document images. The dataset is constructed through an automatic annotation pipeline which leverages a perplexity-based scoring function to identify the document regions most semantically aligned with each answer sentence. At its core, \datasetlong enforces a structural annotation schema consisting of: (i) the document image, (ii) a query, (iii) an answer grounded in the document context, (iv) the spatial coordinates of the document regions supporting the answer, and (v) the semantic type of each evidence region, covering fine-grained categories such as \textit{paragraph/body}, \textit{caption}, \textit{heading/title}, \textit{subtitle/byline}, \textit{data}, \textit{sub-data}, \textit{table}, \textit{image}, \textit{picture}, \textit{list}, \textit{text}, and \textit{other}.

Overall, \datasetlong comprises 237,418 documents spanning diverse domains, including scientific articles, business reports, and digital slides, totaling 295,554 examples. The dataset covers a wide variety of visual elements and content types, with a Simpson coefficient of 0.97 on the images, compared to 0.82 for DUDE when computed under the same setting, highlighting its high visual diversity. It is split into training and test sets with no overlapping images between partitions. The large-scale training set enables effective fine-tuning of MLLMs, demonstrating the effectiveness of the proposed pipeline in generating high-quality supervision signals. The test set serves as a standardized benchmark to evaluate existing and future MLLMs with grounding capabilities on \task.

\subsection{Mask-based Perplexity-Derived Attribution}\label{sec:mappet}
To automatically extract region-answer associations, we propose \ours, which is based on a perplexity-derived score designed to quantify the contribution of visual regions to a language model confidence in generating an answer. Perplexity~\cite{jelinek1977perplexity} measures the probability of generating a sentence through an LLM. Let $T_n=\{t_1, t_2, ..., t_{n-1}\}$ denote a set of answer tokens and $I=\{i_i, i_2, ..., i_m\}$ be the set of image pixels. Given an MLLM, $P(t_k|T_k, I)$ is the conditional probability of generating token $t_k$ given the previous tokens and the input image $I$. For simplicity, we omit the question and other context tokens in the formulation. The image-conditioned perplexity for the answer tokens is defined as
\begin{equation}
    \rho_{I} = \exp \Biggl( - \frac{1}{N} \sum_{i=1}^N \log P(t_i | T_i, I) \Biggr),
\end{equation}
where lower perplexity is associated with higher model confidence in generating the answer.

Let $Q \subset I$ be a region of pixels from $I$ and let $I_Q = I/Q$ be the masked variant of $I$, obtained by occluding the pixels in $Q$. The attribution score of the region $Q$ is defined as
\begin{equation}
    \Delta_{Q} = \log(\rho_{I_{Q}}) - \log (\rho_{I}).
\end{equation}

Intuitively, if the region $Q$ contains information essential for generating the answer, masking it increases the model perplexity, resulting in a positive attribution score. Conversely, regions that are irrelevant or uninformative will have scores near zero or negative, indicating little or no contribution.
By computing $\Delta_Q$ for all candidate regions extracted from the document, \ours identifies the region most responsible for the answer.

\subsection{Dataset Collection}\label{sec:dataset_collection}
To construct \dataset, we aggregate data from eight publicly available document understanding benchmarks. The selected datasets satisfy two criteria aligned with our benchmark objective. First, all datasets provide document-oriented content, consisting of either scanned/rendered PDFs or document page images. This ensures domain consistency and focuses the benchmark on visual-text reasoning within structured documents. Second, each dataset adopts a question-answering format and contains single-page instances or multi-page examples with page-answer alignment, which makes spatial grounding possible.

Specifically, \datasetlong is built from these public datasets: DoclingMatix~\cite{nassar2025smoldocling}, DocVQA~\cite{Mathew_2021_WACV}, VisualMRC~\cite{tanaka2021visualmrc}, LongDocURL~\cite{deng-etal-2025-longdocurl}, MMLongBenchDoc~\cite{ma2024mmlongbench}, SlideVQA~\cite{tanaka2023slidevqa}, VisualWebBench~\cite{liu2024visualwebbench}, and VISA~\cite{ma-etal-2025-visa}, which itself includes three sub-datasets: Wiki-VISA, Paper-VISA, and FineWeb-VISA. Among these, VISA, VisualMRC, and LongDocURL provide existing grounding annotations. We reprocess these samples using our automatic annotation pipeline and retain only those whose attribution score exceeds a predefined threshold. This filtering procedure ensures precise and unambiguous grounding annotations suitable for reliable training and evaluation. 

DoclingMatix and FineWeb-VISA lack official test sets and are therefore used only for training. In contrast, LongDocURL, MMLongBenchDoc, and VisualWebBench are evaluation-oriented benchmarks and are used exclusively for testing. The remaining datasets follow their original train/test splits. 

\begin{wrapfigure}{r}{0.50\linewidth}
\vspace{-0.55cm}
\small
\setlength{\tabcolsep}{0.35em}

\begin{minipage}{1\linewidth}
\centering

\resizebox{\linewidth}{!}{%
\begin{tabular}{lc cc c cc}
     \toprule
    & & \multicolumn{2}{c}{\textbf{Train}} & & \multicolumn{2}{c}{\textbf{Test}} \\
    \cmidrule(lr){3-4} \cmidrule(lr){6-7}
    \textbf{Source} & & \#Docs & \#Q\&A & & \#Docs & \#Q\&A \\
    \midrule
    DoclingMatix~\cite{nassar2025smoldocling}& & 99,816& 139,632& & -& -\\
    DocVQA~\cite{Mathew_2021_WACV}& & 1,114& 4,070& & 123
& 467\\
    VisualMRC~\cite{tanaka2021visualmrc}& & 7,641& 17,068& & 2,143& 4,857\\
    VISA~\cite{ma-etal-2025-visa}& & 116,205& 116,205& & 2,779& 2,779\\
    SlideVQA~\cite{tanaka2023slidevqa}& & 5,567& 8,044& & 981
& 1,235\\
    VisualWebBench~\cite{liu2024visualwebbench}& & -& -& & 108
& 233\\
    LongDocURL~\cite{deng-etal-2025-longdocurl}& & -& -& & 688
& 688\\
    MMLongBench-Doc~\cite{ma2024mmlongbench}& & -& -& & 253& 276\\
    \midrule
    \rowcolor{myhighlight} 
    \textbf{\datasetlong (\dataset)} & & 230,343& 285,019& & 7,075& 10,535\\ 
    \bottomrule

\end{tabular}
}
\vspace{0.2cm}
\manualtablecaption{\datasetlong composition.}
\label{tab:dataset}

\vspace{0.175cm}

\begin{minipage}{0.49\linewidth}
    \centering
    \includegraphics[width=\linewidth]{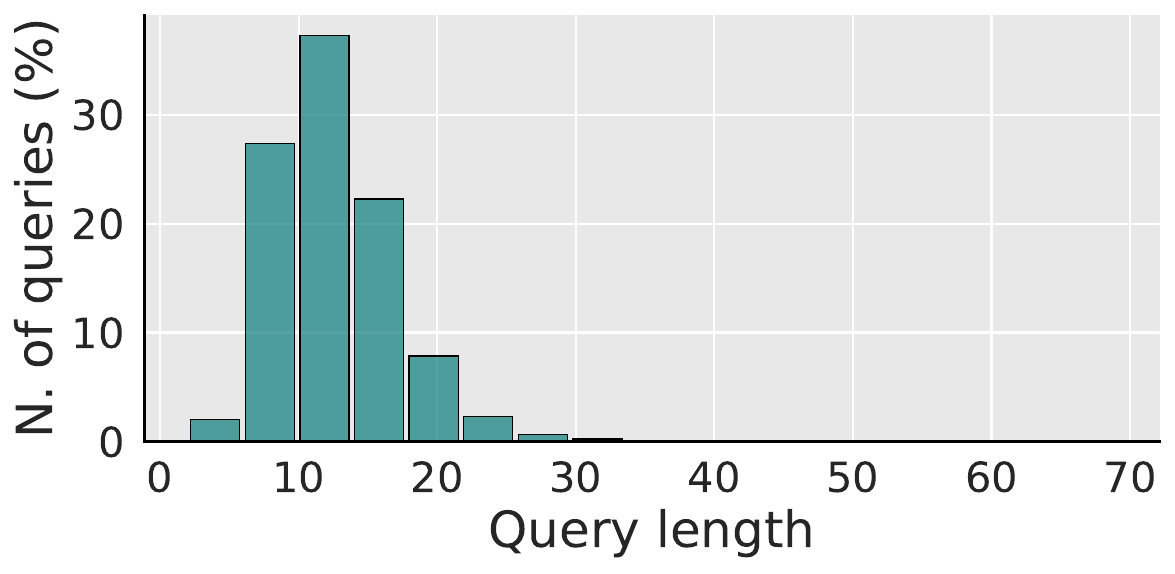}
\end{minipage}
\hfill
\begin{minipage}{0.49\linewidth}
    \centering
    \includegraphics[width=\linewidth]{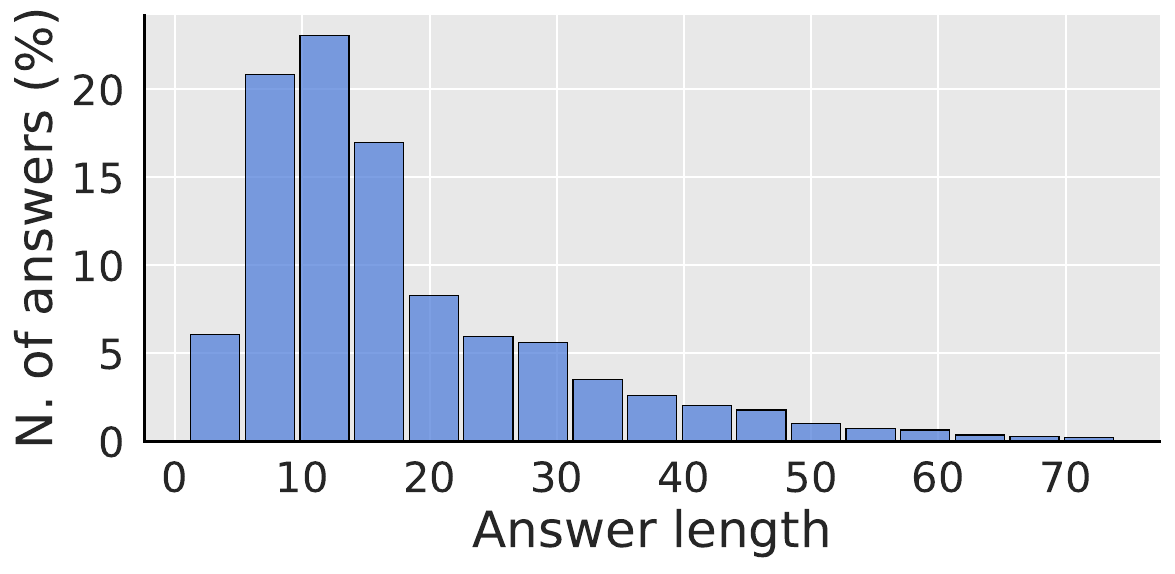}
\end{minipage}
\vspace{-0.2cm}
\caption{Statistics of \datasetlong in terms of query length distribution (left) and answer length distribution (right).}
\vspace{-0.4cm}
\label{fig:docattribench_histograms}
\end{minipage}
\end{wrapfigure}

Complete statistics and split details are presented in Table~\ref{tab:dataset}. Figure~\ref{fig:docattribench_histograms} shows the distributions of query and answer lengths in tokens. Queries are typically short and concentrated around 5--20 tokens, reflecting concise information-seeking questions. Answers exhibit greater variability, with many short responses but also a long tail extending beyond 50 tokens, indicating that DocAttriBench covers both extractive-style answers and more descriptive responses with diverse granularity.

We also analyze the distribution of grounded evidence types in Table~\ref{tab:box_distribution}. Unlike prior document QA benchmarks, which typically provide coarse or text-only grounding, \dataset includes fine-grained bounding boxes across diverse structural categories, including text (paragraphs, headings, captions, lists), tables, and visual elements, among others. These categories are obtained by applying Docling~\cite{Docling} to the document pages, ensuring a consistent layout taxonomy across datasets. This categorization offers a rich resource for studying grounding and reasoning in complex documents.

\begin{figure*}[t]    
  \centering
  \centerline{\includegraphics[width=0.99\linewidth]{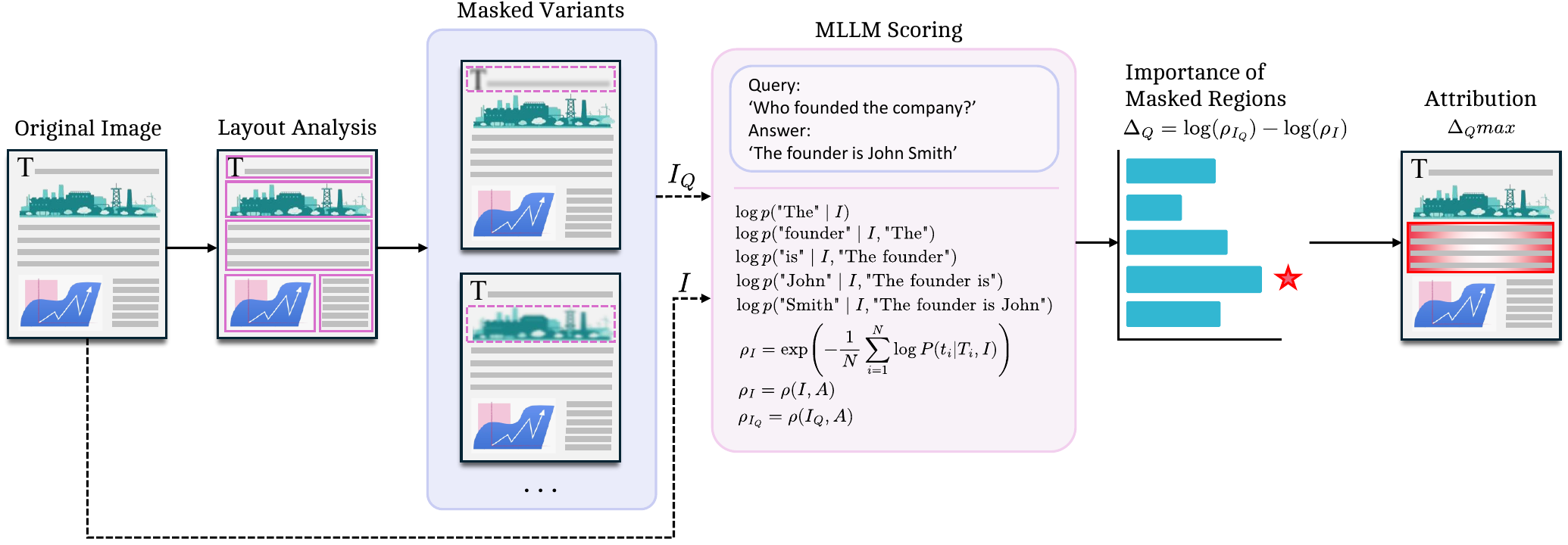}}
\vspace{-0.1cm}
  \caption{Illustration of the \ours visual attribution pipeline.}
  \label{fig:visual_attribution_process}
  \vspace{-0.4cm}
\end{figure*}

\subsection{Automatic Annotation Pipeline}\label{sec:annotation_pipeline}
Our automatic annotation pipeline aims to extract the document regions that are most relevant for the generation of a given answer. First, we filter examples from existing benchmarks, discarding cases with multiple evidence boxes. Second, we extract structural elements from the document pages using a document layout analysis model. In parallel, we abstract existing answers using an MLLM. Then, we use \ours to identify the most relevant structural element with respect to the abstractive answer. Finally, a thresholding mechanism discards examples with low confidence scores and further removes cases where the highest-scoring region is not sufficiently distinguished from the second-best candidate. 
A graphical overview of our automatic annotation pipeline is provided in Figure~\ref{fig:visual_attribution_process}.

\tit{Examples Filtering} To ensure reliable attribution performance, we retain only QA examples linked to a single document page, discarding instances associated with multiple or no pages.

\tit{Document Layout Analysis}
Accurate detection of structural elements is essential for our pipeline, as the dataset must provide consistently sized candidate layout regions for evidence localization. However, datasets lack explicit layout coordinates or have inconsistent annotation granularity. To address this, we apply Docling~\cite{Docling} to the document images; this state-of-the-art off-the-shelf module predicts both bounding boxes and semantic categories of structural elements.

\begin{wraptable}{r}{0.35\textwidth}
\vspace{-0.5cm}
  \centering
  \small
  \setlength{\tabcolsep}{0.35em}
  \resizebox{\linewidth}{!}{%
  \begin{tabular}{lc cc}
    \toprule
    & & \multicolumn{2}{c}{\textbf{Split}} \\
    \cmidrule(lr){3-4}
    \textbf{Box Type} & & \textbf{Train} & \textbf{Test} \\
    \midrule
    Paragraph/Body  & & 148,242 & 5,265 \\
    Caption         & & 15,791  & 347   \\
    Heading/Title   & & 9,994   & 232 
    \\
    Subtitle/Byline & & 2,960   & 189
    \\
    Data            & & 1,342   & 68    \\
    Sub-data        & & 45      & 15
    \\
    Image           & & 835     & 183    \\
    Picture         & & 11,531  & 1,193  \\
    Table           & & 26,893  & 1,175 \\
    List            & & 6,011   & 201
    \\
    Text            & & 6,763   & 1,154 \\
    Other           & & 54,663  & 528
    \\
    \midrule
    \textbf{Total}  & & 285,070 & 10,550 \\
    \bottomrule
  \end{tabular}
  }
  \vspace{-0.15cm}
  \caption{Distribution of answer-evidence bounding box types across dataset splits.}\label{tab:box_distribution}
  \vspace{-0.55cm}
\end{wraptable}

We retain all categories predicted by Docling to preserve fine-grained information, including \textit{paragraph/body}, \textit{caption}, \textit{heading/title},  \textit{subtitle/byline}, \textit{data} (standalone non-tabular numeric content), \textit{sub-data} (secondary data linked to a main content element), \textit{image} (generic visual elements such as logos), \textit{picture} (meaningful visual content such as photos, diagrams, charts, or illustrations), \textit{table} (structured or unstructured tabular content), \textit{list}, generic \textit{text}, and \textit{other}. For VisualMRC, where overlapping annotations are present for reducible elements (\eg~table cells and nested tables), we preserve only the outermost bounding box to maintain annotation coherence.

\tit{Answer Abstraction} To ensure compatibility with generative MLLMs and enable sentence-level visual grounding, we perform answer abstraction to harmonize answer style and granularity across datasets. LLMs are inherently proficient in generating discursive answers rather than extractive ones. However, among the selected sources, most contain extractive answers, which are misaligned with the expressive style expected from LLMs. To harmonize these formats, we employ Qwen2.5-VL-7B~\cite{bai2025qwen25vltechnicalreport}, instructing it to abstract otherwise extractive answers while preserving their semantic content. Model selection criteria and the prompt template for answer abstraction are available in the supplementary material.

\tit{Attribution} The third stage of our pipeline is attribution, which locates the visual evidence supporting each answer. This step is pivotal for constructing a comprehensive dataset suitable for grounding evaluation. Specifically, for every example in the benchmark dataset and for each sentence in the abstracted answers, we use MAPPET to compute an attribution score for every structural element. The elements are then ranked in descending order of relevance and the top-ranked element is selected as the supporting evidence.

\tit{Filtering} As a final step, we filter the examples where the system fails to identify a reliable attribution. This step addresses three primary cases. First, some answers might correspond to information redundantly appearing in multiple locations within the page, preventing unique grounding. Second, some sentences might depend on evidence scattered across several regions. Third, some answers might not be answerable with the document page. 

To handle these scenarios, we employ a two-step filtering strategy. We first discard all samples whose top attribution score is lower than a fixed confidence threshold. Then, we remove cases where the score gap between the two highest-ranked regions is smaller than a predefined margin. For benchmarks already containing the evidence source, we compute the region attribution score and retain the example if the score is greater than the threshold. This strategy ensures that the retained examples exhibit strong evidence alignment and are not ambiguous or attributable to multiple sources. For all the aforementioned cases, we set the threshold to the same value ($\tau = 0.1$).

\section{Evaluation Protocol\vspace{-0.1cm}}
It is challenging to evaluate whether MLLMs correctly ground their answers, because they generate natural, free-form text rather than fixed, easily comparable outputs.
Our evaluation protocol is designed to satisfy two key requirements: it must enable the assessment of free-form text, and it must jointly evaluate both answer generation and grounding. To this end, we design an evaluation protocol on \datasetlong to assess MLLMs on three complementary tasks: (i) grounded answer generation, where the model receives a query and document image, and must generate both a textual answer and its visual grounding, (ii) post-hoc grounding, where the model is given a query, document image, and the reference answer, and must predict only the bounding box of the supporting evidence, and (iii) answer-locating, where the model receives a query and document image only, and must output the coordinates of the evidence without generating any textual answer. All our prompt templates are available in the supplementary material. 

\subsection{Evaluation Metrics}\label{sec:eval_metr}
\tinytit{Answer Accuracy}
Current MLLMs are optimized for generating long-form, open-ended answers rather than concise, extractive responses. To fairly evaluate them, our protocol accounts for differences in answer style and length across models. In particular, we build on prior work from MATHVISTA~\cite{lu2023mathvista} and MMLongBenchDoc~\cite{ma2024mmlongbench}. Specifically, we repurpose the LLM-based answer extraction module from the latter, applying it to break each generated answer into a structured list of information units, where each unit represents a distinct piece of information. To support further processing, the model is also instructed to assign a semantic type to each extracted unit.

During evaluation, we align these extracted units with the ground-truth answer spans. For text, we use decreasing Average Normalized Levenshtein Similarity (ANLS)~\cite{biten2019scene}, while for numeric values we require an exact match. A text span pair is considered a match if its similarity exceeds a fixed threshold ($\tau = 0.5$). Finally, we compute answer-level accuracy as the proportion of ground-truth spans that have a corresponding match in the generated answer.

\tit{Box F1} MLLMs with grounding capabilities produce bounding boxes in diverse formatting conventions. Our evaluation protocol accounts for this variability with adaptable answer-generation prompts and model-specific parsing functions, which are used to accurately extract the predicted boxes. Ground truth and predicted regions are matched by descending order of Intersection over Union (IoU) and are considered correct matches if their IoU exceeds a fixed threshold ($\tau = 0.5$). Unmatched ground truth boxes are accounted as false negatives, while unmatched predictions are counted as false positives. Finally, we compute the F1 score to quantify grounding accuracy.

\tit{Overall Answer Correctness} To fully assess the correctness of the generated answers, we combine Answer Accuracy and BoxF1 into an overall answer-level correctness metric. For a given answer, the Overall Answer Correctness is 1 if both Answer Accuracy and BoxF1 exceed a threshold ($\tau = 0.5$).

\subsection{Evaluating MLLMs on \datasetlong}\label{sec:eval_mllm}
\tinytit{Zero-shot MLLMs} To establish a comprehensive zero-shot benchmark on our dataset, we evaluate sixteen representative MLLMs with localization capabilities. From the Qwen family, we include Qwen2.5-VL~\cite{bai2025qwen25vltechnicalreport} (in the 3B, 7B, and 32B sizes) and Qwen3-VL~\cite{yang2025qwen3technicalreport} (in the 2B, 8B, and 32B variants). From the InternVL series, we select InternVL2.5~\cite{chen2024expanding}, InternVL3~\cite{zhu2025internvl3exploringadvancedtraining}, and InternVL3.5~\cite{wang2025internvl35advancingopensourcemultimodal}, each evaluated in 2B, 8B, and 38B variants. Finally, we include VISA~\cite{ma-etal-2025-visa}, a document \task oriented fine-tuned derivative of Qwen2-VL~\cite{wang2024qwen2}.

\tit{Fine-tuning MLLMs} To evaluate the effectiveness of the constructed dataset for training grounding-capable MLLMs, we fine-tune Qwen2.5-VL-7B\footnote{\href{https://huggingface.co/Qwen/Qwen2.5-VL-7B-Instruct}{\texttt{Qwen/Qwen2.5-VL-7B-Instruct}}}, Qwen3-VL-8B\footnote{\href{https://huggingface.co/Qwen/Qwen3-VL-8B-Instruct}{\texttt{Qwen/Qwen3-VL-8B-Instruct}}}, and InternVL3.5-8B\footnote{\href{https://huggingface.co/OpenGVLab/InternVL3_5-8B-Instruct}{\texttt{OpenGVLab/InternVL3-5-8B-Instruct}}} using Low-Rank Adaptation (LoRA)~\cite{hu2022lora} as implemented in the PEFT library~\cite{peft}. LoRA is applied to attention and MLP layers, with $r{=}16$, $\alpha{=}32$, and $0.05$ dropout. While Qwen2.5-VL-7B is used in our pipeline for perplexity-based filtering and answer abstraction, we fine-tune InternVL3.5-8B to assess the dataset effectiveness in training a model outside the Qwen family. This mitigates the risk of architecture-specific bias and ensures that observed performance gains reflect improved grounding capabilities rather than adaptation to Qwen-specific annotation patterns.

For fine-tuning, we employ a modified version of the original inference prompt templates, removing bounding-box dimensions instructions to let the model internalize the task structure through data exposure. For each training sample, a task type is randomly selected from the three defined evaluation tasks to ensure balanced task diversity. This prevents the model from overfitting to a single prompting pattern and encourages it to handle multiple grounding tasks requiring different output formats. Moreover, the image size is randomized to ensure robustness with differing image dimensions. For grounding supervision, bounding boxes are represented in absolute coordinates for Qwen2.5 and relative coordinates ([0, 1000]) for both Qwen3 and InternVL3.5, following their pretraining conventions. Training is conducted for 24 hours on two A100 GPUs with learning rate of 1e-4 and a batch size of 32. We name the models fine-tuned from Qwen2.5 and Qwen3 as \vsa-7B and \vsa-8B, respectively, and the model fine-tuned from InternVL3.5 as \vsa-I-8B.

\begin{table*}[t]
  \centering
  \small
  \setlength{\tabcolsep}{0.28em}
  \resizebox{\linewidth}{!}{%
  \begin{tabular}{lc ccc ccc ccc ccc ccc ccc ccc}
    \toprule
    & & 
    \multicolumn{3}{c}{\textbf{DocVQA}} &
    \multicolumn{3}{c}{\textbf{VisualMRC}} &
    \multicolumn{3}{c}{\textbf{VISA}} &
    \multicolumn{3}{c}{\textbf{SlideVQA}} &
    \multicolumn{3}{c}{\textbf{VisualWebB}} &
    \multicolumn{3}{c}{\textbf{LongDocURL}} &
    \multicolumn{3}{c}{\textbf{MMLongB-Doc}}  \\
    \cmidrule(lr){3-5}  \cmidrule(lr){6-8}  \cmidrule(lr){9-11}  \cmidrule(lr){12-14}  \cmidrule(lr){15-17}  \cmidrule(lr){18-20}  \cmidrule(lr){21-23}
    \textbf{Model} & & $\mathsf{Acc}$ & $\mathsf{Acc}_\text{F}$ & $\mathsf{\% Filt}$ & $\mathsf{Acc}$ & $\mathsf{Acc}_\text{F}$ & $\mathsf{\% Filt}$ & $\mathsf{Acc}$ & $\mathsf{Acc}_\text{F}$ & $\mathsf{\% Filt}$ & $\mathsf{Acc}$ & $\mathsf{Acc}_\text{F}$ & $\mathsf{\% Filt}$ & $\mathsf{Acc}$ & $\mathsf{Acc}_\text{F}$ & $\mathsf{\% Filt}$ & $\mathsf{Acc}$ & $\mathsf{Acc}_\text{F}$ & $\mathsf{\% Filt}$ & $\mathsf{Acc}$ & $\mathsf{Acc}_\text{F}$ & $\mathsf{\% Filt}$ \\
    \midrule
    DeepSeek-VL-7B & & 70.0 & 87.0 & 71.0 & 72.7 & \textbf{94.1} & \textbf{94.1} & 20.7 & 55.1 & 42.9 & 63.6 & 91.9 & 90.3 & 56.6 & 88.9 & 86.7 & 58.0 & 86.0 & 67.4 & 46.8 & 88.6 & 77.1\\
    Qwen3-VL-8B & & 82.0 & 90.7 & 81.4 & 54.5 & 87.7 & 87.7 & 46.0 & 83.7 & 74.8 & 45.5 & 82.2 & 80.0 & 71.7 & \textbf{96.9} & \textbf{92.2} & 65.0 & \textbf{95.8} & \textbf{79.2} & 54.3 & \textbf{95.2} & 85.7\\
    \midrule
    Qwen2.5-VL-7B (ext) & & \textbf{95.0} & 95.8 & 86.5 & \textbf{84.8} & 91.2 & 89 & \textbf{72.5} & 84.4 & 78.7 & 77.1 & 87.5 & 87.5 & \textbf{82.8} & 90.5 & 84.5 & \textbf{83.0} & 90.8 & 71.3 & 63.4 & 82.1 & 76.1\\
    \rowcolor{myhighlight}
    \textbf{Qwen2.5-VL-7B (abs)} & & 93.0 & \textbf{98.9} & \textbf{88.0} & 74.5 & 91.8 & 91.8 & 63.2 & 84.2 & \textbf{79.4} & \textbf{80.6} & \textbf{93.6} & \textbf{92.3} & 78.8 & 94.6 & 87.8 & 70.0 & 95.2 & 74.6 & \textbf{66.0} & 89.7 & \textbf{86.2}\\
    \midrule
    \rowcolor{TitleColor}
    Qwen2.5-VL-32B & & 90.9 & 95.6 & 86.8 & 67.3 & 92.2 & 92.2 & 57.8 & \textbf{88.1} & 78.8 & 63.3 & 91.9 & 90.3 & 66.3 & 89.6 & 83.6 & 78.0 & 97.1 & 78.6 & 61.7 & 89.1 & 78.3 \\
    \bottomrule
  \end{tabular}
    }
    \vspace{-0.15cm}
    \caption{\ours annotation quality. \textit{Acc} denotes overall annotation accuracy; \textit{Acc}$_F$ is accuracy on \ours-filtered annotations; \%\textit{Filt} reports accuracy when hallucinated content is automatically marked incorrect.}
  \label{tab:annotation_results}
  \vspace{-0.4cm}
\end{table*}

\section{Experimental Results}
We report experimental results to assess both benchmark quality of \datasetlong and the attribution performance of current MLLMs. We first assess the reliability of the \ours-generated evidence annotations, including their alignment with human judgments and their sufficiency and necessity for supporting answer generation. We then analyze the zero-shot performance of existing MLLMs on \dataset, followed by attribution-based approaches, including VISA and our three fine-tuned models. The human annotation quality analysis is reported in Table~\ref{tab:annotation_results}, the evidence sufficiency and necessity analysis in Table~\ref{tab:evidence_sufficiency_necessity_per_dataset}, and the main results on the \dataset test set in Tables~\ref{tab:results} and~\ref{tab:results2}.

\subsection{Annotation Quality Evaluation}
To validate \ours's effectiveness in generating a grounding dataset, we compare its machine-generated annotations with human annotations.
We select open-source models, which we use for both generating abstractive answers and producing attributions. Specifically, we evaluate Qwen2.5-VL-7B, Qwen3-VL-8B, DeepSeek-VL-7B and Qwen2.5-VL-32B. The first three models are chosen for their favorable trade-off between computational cost and inference speed, enabling large scale annotation. The 32B variant allows us to examine whether scaling improves attribution quality. Empirically, the 32B model provides inconsistent improvements, so we refrain from testing even larger models in our study.

We randomly sample 100 examples from each source benchmark (including 100 from Paper-VISA and 100 from Wiki-VISA). Three annotators then identify the evidence boxes for each sampled example and mark any form of hallucination in the original answers of the dataset. When multiple valid evidence boxes exist, all are retained. We report annotation quality metrics in Table~\ref{tab:annotation_results}. This evaluation allows us to assess both the raw attribution quality of each model and the effectiveness of the \ours filtering strategy in retaining reliable evidence annotations. Annotation accuracy ($\mathsf{Acc}$) measures the portion of examples where the machine-predicted box exactly matches one of the annotated ground-truth boxes. Across benchmarks, $\mathsf{Acc}$ ranges from 20\% to 95\% due to heterogeneous underlying models. After applying \ours filtering mechanism, we recompute accuracy on the retained subsets ($\mathsf{Acc}_F$) and observe consistent increases (sometimes exceeding 95\%), highlighting \ours's ability to effectively discard noisy and unreliable annotations.

Finally, we treat all hallucinated examples as incorrect, even when the predicted region is semantically consistent with the answer, and compute the final accuracy (\%$\mathsf{Filt}$). This evaluation produces a slight decrease in performance, but it is necessary to highlight the amount of noise present in existing benchmarks. Overall, these results confirm that \ours yields high-quality attributions and that its filtering step plays an important role in enhancing dataset reliability.

For Qwen2.5-VL-7B, we ablate answer abstraction by replacing abstractive answers (abs) with extractive ones (ext). The results are comparable, with a slight advantage for extractive answers; however, after filtering, abstractive answers consistently yield higher attribution accuracy, suggesting more informative signals for \ours filtering.

\begin{table*}[t]
\centering
\vspace{0.3cm}
\resizebox{\linewidth}{!}{
\begin{tabular}{llcccccccc}
\toprule
\textbf{Metric}
& \textbf{Model}
& \textbf{DocVQA}
& \textbf{VisualMRC}
& \textbf{VISA}
& \textbf{SlideVQA}
& \textbf{VisualWebB}
& \textbf{LongDocURL}
& \textbf{MMLongB-Doc}
& \textbf{Overall} \\
\midrule
\multirow{2}{*}{Evidence-only}
& Qwen2.5 & $+0.136$ & $+0.009$ & $+0.174$ & $+0.052$ & $+0.018$ & $-0.014$ & $-0.060$ & $\mathbf{+0.061}$ \\
& InternVL2.5 & $+0.652$ & $+0.053$ & $+0.071$ & $+0.047$ & $+0.037$ & $-0.060$ & $-0.012$ & $\mathbf{+0.107}$ \\
\midrule
\multirow{2}{*}{Evidence-masked}
& Qwen2.5 & $-1.368$ & $-1.094$ & $-0.624$ & $-0.938$ & $-0.971$ & $-0.941$ & $-0.753$ & $\mathbf{-0.914}$ \\
& InternVL2.5 & $-1.013$ & $-0.855$ & $-0.436$ & $-0.643$ & $-0.876$ & $-0.844$ & $-0.607$ & $\mathbf{-0.714}$ \\
\midrule
\multirow{2}{*}{Masked worse (\%)}
& Qwen2.5 & $99.0$ & $99.5$ & $96.8$ & $99.5$ & $99.5$ & $99.5$ & $100.0$ & $\mathbf{98.8}$ \\
& InternVL2.5 & $84.0$ & $92.0$ & $84.3$ & $87.5$ & $96.5$ & $95.5$ & $87.0$ & $\mathbf{88.9}$ \\
\midrule
\multirow{2}{*}{Random-masked}
& Qwen2.5 & $-0.288$ & $-0.362$ & $-0.098$ & $-0.458$ & $-0.148$ & $-0.125$ & $-0.280$ & $\mathbf{-0.232}$ \\
& InternVL2.5 & $-0.198$ & $-0.272$ & $-0.082$ & $-0.314$ & $-0.133$ & $-0.123$ & $-0.216$ & $\mathbf{-0.178}$ \\
\bottomrule
\end{tabular}
}
\vspace{-0.15cm}
\caption{Evidence sufficiency and necessity analysis across test sets. Average token log-likelihood deltas for evidence-only, evidence-masked, and random-masked inputs, together with the percentage of examples where evidence masking reduces answer likelihood.}
\label{tab:evidence_sufficiency_necessity_per_dataset}
\vspace{-0.3cm}
\end{table*}

\begin{table*}[t]
  \centering
  \small
  \setlength{\tabcolsep}{0.2em}
  \vspace{0.15cm}
  \resizebox{\textwidth}{!}{%
  \begin{tabular}{lc ccc ccc ccc ccc ccc ccc ccc cc}
    \toprule
     & & 
    \multicolumn{3}{c}{\textbf{DocVQA}} &
    \multicolumn{3}{c}{\textbf{VisualMRC}} &
    \multicolumn{3}{c}{\textbf{VISA}} &
    \multicolumn{3}{c}{\textbf{SlideVQA}} &
    \multicolumn{3}{c}{\textbf{VisualWebB}} &
    \multicolumn{3}{c}{\textbf{LongDocURL}} &
    \multicolumn{3}{c}{\textbf{MMLongB-Doc}}  \\
    \cmidrule(lr){3-5}  \cmidrule(lr){6-8}  \cmidrule(lr){9-11}  \cmidrule(lr){12-14}  \cmidrule(lr){15-17}  \cmidrule(lr){18-20}  \cmidrule(lr){21-23}
    \textbf{Model} & & $\mathsf{Acc}_\text{txt}$ & $\mathsf{F1}_\text{box}$ & $\mathsf{Acc}$ & $\mathsf{Acc}_\text{txt}$ & $\mathsf{F1}_\text{box}$ & $\mathsf{Acc}$ & $\mathsf{Acc}_\text{txt}$ & $\mathsf{F1}_\text{box}$ & $\mathsf{Acc}$ & $\mathsf{Acc}_\text{txt}$ & $\mathsf{F1}_\text{box}$ & $\mathsf{Acc}$ & $\mathsf{Acc}_\text{txt}$ & $\mathsf{F1}_\text{box}$ & $\mathsf{Acc}$ & $\mathsf{Acc}_\text{txt}$ & $\mathsf{F1}_\text{box}$ & $\mathsf{Acc}$ & $\mathsf{Acc}_\text{txt}$ & $\mathsf{F1}_\text{box}$ & $\mathsf{Acc}$ & & \textbf{Avg} \\
    \midrule
    \rowcolor{TitleColor}
    \multicolumn{25}{l}{\textit{Zero-shot MLLMs}} \\
    InternVL2.5-2B & & 37.5 & 0.0 & 0.0 & 30.0 & 1.0 & 0.1 & 6.6 & 0.0 & 0.0 & 29.6 & 0.7 & 0.0 & 33.5 & 0.5 & 0.4 & 20.1 & 0.2 & 0.0 & 25.7 & 0.4 & 0.0 & & 0.1 \\
    InternVL3-2B & & 41.9 & 4.9 & 0.4 & 25.9 & 8.2 & 2.1 & 10.4 & 0.0 & 0.0 & 14.9 & 23.2 & 1.6 & 37.3 & 3.5 & 1.3 & 16.7 & 4.7 & 0.0 & 15.8 & 8.5 & 0.7 & & 0.7\\
    InternVL3.5-2B & & 49.0 & 19.3 & 11.7 & 23.1 & 2.0 & 0.9 & 2.3 & 0.4 & 0.0 & 39.6 & 0.6 & 0.3 & 18.4 & 9.8 & 3.9 & 22.2 & 1.0 & 0.6 & 18.8 & 1.5 & 0.7 & & 2.1\\
    Qwen2.5-VL-3B & & 12.4 & 4.3 & 0.2 & 34.3 & 11.5 & 2.7 & 2.7 & 0.1 & 0.0 & 20.6 & 17.6 & 3.1 & 6.0 & 7.1 & 0.4 & 12.7 & 6.1 & 0.6 & 11.8 & 11.6 & 0.0 & & 0.9\\
    Qwen3-VL-2B & & 29.9 & 6.8 & 6.3 & 34.7 & 17.0 & 9.1 & 27.8 & 0.4 & 0.1 & 29.9 & 10.4 & 7.2 & 33.5 & 5.1 & 3.4 & 14.4 & 7.2 & 2.8 & 15.8 & 8.1 & 5.2 & & 4.3\\
    \midrule
    InternVL2.5-8B & & 59.0 & 0.0 & 0.0 & 65.7 & 2.8 & 1.2 & 36.0 & 0.0 & 0.0 & 64.8 & 2.8 & 1.7 & 60.9 & 0.9 & 0.4 & 36.6 & 0.0 & 0.0 & 45.6 & 1.2 & 0.7 & & 0.5\\
    InternVL3-8B & & 64.9 & 6.4 & 3.2 & 63.9 & 8.2 & 4.2 & 21.9 & 0.0 & 0.0 & 61.9 & 16.5 & 9.3 & 53.2 & 1.9 & 1.3 & 40.7 & 3.9 & 1.9 & 44.1 & 5.4 & 1.8 & & 2.6\\
    InternVL3.5-8B & & 70.3 & 22.0 & 16.9 & 68.1 & 5.4 & 2.9 & 37.6 & 0.4 & 0.2 & 70.5 & 4.0 & 2.2 & 64.0 & 25.0 & 17.6 & 45.5 & 3.2 & 1.6 & 58.5 & 2.1 & 1.5 & & 5.3\\
    Qwen2.5-VL-7B & & 67.0 & 7.3 & 2.6 & 65.6 & 10.0 & 2.7 & 52.5 & 0.6 & 0.0 & 65.1 & 17.8 & 8.1 & 61.4 & 7.7 & 3.0 & 44.6 & 10.7 & 3.7 & 51.5 & 12.5 & 3.7 & & 3.0\\
    Qwen3-VL-8B & & 67.7 & 14.6 & 10.8 & 67.8 & 24.4 & 14.3 & 53.1 & 4.2 & 1.9 & 70.1 & 9.4 & 7.0 & 63.1 & 16.8 & 8.6 & 43.1 & 19.8 & 7.0 & 52.2 & 14.9 & 7.7 & & 9.4\\
    \midrule
    InternVL2.5-38B & & 64.2 & 1.0 & 0.4 & 65.9 & 6.9 & 3.8 & 5.8 & 0.3 & 0.1 & 63.0 & 10.0 & 6.3 & 63.1 & 2.8 & 1.7 & 40.1 & 3.0 & 1.3 & 48.9 & 6.0 & 3.7 & & 2.3\\
    InternVL3-38B & & 59.0 & 4.6 & 3.2 & 66.9 & 6.4 & 3.9 & 44.9 & 1.8 & 0.4 & 66.9 & 11.4 & 8.3 & 60.1 & 2.1 & 1.7 & 38.4 & 4.0 & 2.2 & 55.1 & 6.8 & 4.4 & & 3.2\\
    InternVL3.5-38B & & 69.0 & 26.4 & 20.4 & 67.2 & 14.9 & 8.6 & 42.4 & 1.4 & 0.2 & 69.2 & 6.2 & 4.3 & 65.2 & 30.0 & 20.2 & 42.4 & 11.5 & 4.9 & 56.2 & 7.6 & 5.2 & & 8.6\\
    Qwen2.5-VL-32B & & 68.1 & 9.7 & 7.8 & 67.3 & 19.6 & 11.6 & 56.4 & 1.4 & 0.6 & 69.5 & 11.5 & 8.9 & 64.0 & 28.4 & 19.3 & 45.9 & 9.8 & 5.8 & 56.6 & 14.8 & 11.8 & & 9.1\\
    Qwen3-VL-32B & & \textbf{72.9} & 41.8 & 29.3 & 68.2 & 56.7 & 36.6 & \textbf{56.5} & 31.0 & 18.3 & \textbf{73.2} & 30.6 & 22.4 & 63.5 & 31.2 & 21.0 & 51.6 & 47.4 & 24.9 & \textbf{61.8} & 44.3 & \textbf{29.0} & & 25.4\\
    \midrule
    \rowcolor{TitleColor}
    \multicolumn{25}{l}{\textit{Source Attribution MLLMs}} \\
    VISA-7B & & 60.5 & 43.2 & 22.1 & 62.6 & 64.3 & 36.6 & 52.2 & 50.4 & 27.1 & 62.2 & 62.6 & 34.1 & 53.6 & 14.5 & 8.6 & 44.1 & 43.0 & 16.4 & 46.0 & 55.9 & 21.3 & & 24.3 \\
    \rowcolor{myhighlight} 
    \textbf{\vsa-7B} & & 66.6 & 65.8 & 40.2 & 73.9 & 74.5 & 52.3 & 46.6 & 58.5 & 29.4 & 66.9 & 65.9 & 39.9 & \textbf{73.9} & 30.7 & 19.6 & \textbf{51.7} & 59.8 & 28.1 & 50.6 & 60.3 & 27.2 & & 33.8 \\
    \rowcolor{myhighlight}
    \textbf{\vsa-8B} & & 61.9 & \textbf{75.3} & \textbf{44.1} & \textbf{74.1} & \textbf{76.7} & \textbf{54.5} & 43.8 & 60.7 & \textbf{29.9} & 66.1 & \textbf{73.3} & \textbf{44.4} & 69.1 & 36.8 & \textbf{26.8} & 50.3 & \textbf{69.4} & \textbf{31.8} & 44.4 & \textbf{66.7} & 27.6 & & \textbf{37.0}\\
    \rowcolor{myhighlight}
    \textbf{\vsa-I-8B} & & 51.4 & 69.8 & 33.7 & 74.0 & 76.0 & 53.2 & 42.3 & \textbf{61.2} & 26.6 & 67.8 & 65.3 & 38.7 & 61.3 & \textbf{40.4} & 25.3 & 48.5 & 68.0 & 29.3 & 47.3 & 66.5 & 26.8 & & 33.4\\
    \bottomrule
  \end{tabular}
  }
   \vspace{-0.15cm}
    \caption{Evaluation of selected models on \datasetlong for grounded answer generation. $\mathsf{Acc}_\text{txt}$: answer accuracy; $\mathsf{F1}_\text{box}$: grounding F1 score; $\mathsf{Acc}$: overall answer correctness.}
    \label{tab:results}
  \vspace{-0.4cm}
\end{table*}

\subsection{Evidence Sufficiency and Necessity}
Beyond comparing machine-generated annotations with human judgments, we assess whether the evidence regions selected from \dataset are sufficient and necessary for answer generation. We compute the average token log-likelihood of the answer on 200 samples from each test set and 400 samples from VISA to cover its two subsets (1,600 examples total), using both Qwen2.5-VL-7B, employed in the \ours pipeline, and the independent InternVL2.5-8B.

For each example, we compare the original full-page image with three interventions. In the \emph{evidence-only} setting, all content except the selected evidence is masked, testing sufficiency: the attributed region alone should preserve the information required for the answer. In the \emph{evidence-masked} setting, only the selected evidence is masked, testing necessity: removing it should reduce answer likelihood. As a control, \emph{random-masked} masks a random region of the same size. We report average token log-likelihood deltas relative to the full-page input, expected to be near zero for evidence-only and strongly negative for evidence-masked, together with \emph{masked worse}, the percentage of cases where evidence masking lowers likelihood.

As shown in Table~\ref{tab:evidence_sufficiency_necessity_per_dataset}, the results are consistent across both models. With Qwen2.5-VL-7B, evidence-only yields an overall delta of $+0.061$, indicating that the selected evidence preserves, and may even improve, answer likelihood relative to the full-page input, while masking the selected evidence produces a much larger negative change of $-0.914$ and lowers likelihood in $98.8\%$ of examples. InternVL2.5-8B independently confirms this trend: evidence-only yields an even larger positive delta of $+0.107$, whereas evidence masking gives $-0.714$ and worsens likelihood in $88.9\%$ of cases. Importantly, masking a random region of the same size has a substantially smaller effect ($-0.232$ for Qwen2.5-VL-7B and $-0.178$ for InternVL2.5-8B). These results indicate that the selected regions are both sufficient and necessary for answer generation, with consistent behavior across both MLLMs.

\begin{table*}[t]
  \centering
  \small
  \setlength{\tabcolsep}{0.38em}
  \vspace{0.15cm}
  \resizebox{\textwidth}{!}{%
  \begin{tabular}{lc cc cc cc cc cc cc cc ccc}
    \toprule
     & & 
    \multicolumn{2}{c}{\textbf{DocVQA}} &
    \multicolumn{2}{c}{\textbf{VisualMRC}} &
    \multicolumn{2}{c}{\textbf{VISA}} &
    \multicolumn{2}{c}{\textbf{SlideVQA}} &
    \multicolumn{2}{c}{\textbf{VisualWebB}} &
    \multicolumn{2}{c}{\textbf{LongDocURL}} &
    \multicolumn{2}{c}{\textbf{MMLongB-Doc}}  \\
    \cmidrule(lr){3-4}  \cmidrule(lr){5-6}  \cmidrule(lr){7-8}  \cmidrule(lr){9-10}  \cmidrule(lr){11-12}  \cmidrule(lr){13-14}  \cmidrule(lr){15-16}
    \textbf{Model} & & $\mathsf{F1}^\text{Q}_\text{box}$ & $\mathsf{F1}^\text{QA}_\text{box}$  & $\mathsf{F1}^\text{Q}_\text{box}$ & $\mathsf{F1}^\text{QA}_\text{box}$ & $\mathsf{F1}^\text{Q}_\text{box}$ & $\mathsf{F1}^\text{QA}_\text{box}$ & $\mathsf{F1}^\text{Q}_\text{box}$ & $\mathsf{F1}^\text{QA}_\text{box}$ & $\mathsf{F1}^\text{Q}_\text{box}$ & $\mathsf{F1}^\text{QA}_\text{box}$ & $\mathsf{F1}^\text{Q}_\text{box}$ & $\mathsf{F1}^\text{QA}_\text{box}$ & $\mathsf{F1}^\text{Q}_\text{box}$ & $\mathsf{F1}^\text{QA}_\text{box}$ & & \textbf{Avg\textsuperscript{Q}} & \textbf{\textbf{Avg\textsuperscript{QA}}} \\
    \midrule
    \rowcolor{TitleColor}
    \multicolumn{19}{l}{\textit{Zero-shot MLLMs}} \\
    InternVL2.5-2B & & 0.0 & 0.2 & 1.1 & 0.4 & 0.2 & 0.0 & 0.6 & 0.3 & 0.0 & 0.0 & 0.2 & 0.2 & 0.5 & 0.0 & & 0.4 & 0.2 \\
    InternVL3-2B & & 8.0 & 15.7 & 12.2 & 13.1 & 2.3 & 4.5 & 13.8 & 21.0 & 4.3 & 6.4 & 3.4 & 5.6 & 6.6 & 12.0 & & 7.2 & 11.2 \\
    InternVL3.5-2B & & 11.1 & 13.8 & 0.8 & 2.0 & 0.2 & 0.2 & 0.2 & 0.3 & 4.8 & 11.7 & 0.2 & 1.2 & 1.2 & 2.5 & & 2.6 & 4.5 \\
    Qwen2.5-VL-3B & & 6.3 & 4.5 & 14.7 & 14.4 & 6.0 & 7.2 & 25.4 & 22.6 & 8.1 & 8.2 & 8.2 & 11.0 & 13.5 & 17.3 & & 11.7 & 12.2 \\
    Qwen3-VL-2B & & 28.2 & 26.4 & 27.7 & 30.4 & 16.5 & 11.4 & 30.9 & 32.7 & 15.9 & 16.5 & 28.0 & 26.4 & 29.6 & 32.5 & & 25.3 & 26.1 \\
    \midrule
    InternVL2.5-8B & & 0.22 & 0.6 & 7.5 & 7.7 & 0.7 & 1.0 & 4.0 & 4.2 & 0.9 & 0.9 & 0.6 & 1.3 & 1.1 & 2.9 & & 2.1 & 2.7 \\
    InternVL3-8B & & 14.6 & 15.2 & 9.9 & 11.1 & 3.6 & 1.8 & 23.2 & 16.0 & 8.4 & 11.6 & 8.0 & 5.7 & 10.3 & 8.1 & & 11.1 & 9.9 \\
    InternVL3.5-8B & & 29.3 & 44.7 & 10.1 & 9.7 & 3.8 & 6.1 & 17.5 & 15.1 & 24.9 & 27.7 & 9.8 & 11.4 & 11.9 & 14.0 & & 15.3 & 18.4 \\
    Qwen2.5-VL-7B & & 16.8 & 19.8 & 22.1 & 25.9 & 10.6 & 10.9 & 25.8 & 32.0 & 13.0 & 17.6 & 16.8 & 20.4 & 20.1 & 27.6 & & 17.9 & 22.0 \\
    Qwen3-VL-8B & & 17.6 & 27.4 & 31.4 & 41.3 & 21.3 & 24.4 & 12.4 & 18.7 & 15.9 & 24.4 & 26.5 & 35.5 & 22.8 & 31.8 & & 21.1 & 29.1 \\
    \midrule
    InternVL2.5-38B & & 10.7 & 11.4 & 16.3 & 15.6 & 6.8 & 6.3 & 27.7 & 27.8 & 6.4 & 6.8 & 12.8 & 13.5 & 29.5 & 27.0 & & 15.8 & 15.5 \\
    InternVL3-38B & & 19.3 & 17.8 & 16.4 & 14.7 & 9.1 & 8.3 & 37.8 & 33.3 & 3.9 & 4.3 & 16.9 & 17.5 & 26.6 & 27.2 & & 18.6 & 17.6 \\
    InternVL3.5-38B & & 28.6 & 36.2 & 25.1 & 23.2 & 4.7 & 4.2 & 10.4 & 11.9 & 28.0 & 30.7 & 14.0 & 13.4 & 12.1 & 13.9 & & 17.6 & 19.1 \\
    Qwen2.5-VL-32B & & 14.8 & 19.1 & 27.7 & 34.6 & 5.12 & 7.2 & 20.4 & 27.4 & 26.9 & 28.8 & 17.9 & 29.2 & 22.4 & 32.4 & & 19.3 & 25.5 \\
    Qwen3-VL-32B & & 47.5 & 51.0 & 61.0 & 64.6 & 38.9 & 42.7 & 35.1 & 39.5 & 28.2 & 30.7 & 44.2 & 46.9 & 54.6 & 51.9 & & 44.2 & 46.8 \\
    \midrule
    \rowcolor{TitleColor}
    \multicolumn{19}{l}{\textit{Source Attribution MLLMs}} \\
    VISA-7B & & 41.3 & 44.5 & 65.4 & 65.8 & \textbf{66.2} & \textbf{73.5} & 62.7 & 62.8 & 13.7 & 14.5 & 42.7 & 46.2 & 55.1 & 56.5 & & 49.6 & 52.0 \\
    \rowcolor{myhighlight} 
    \textbf{\vsa-7B} & & 57.6 & 60.8 & 64.9 & 68.6 & 36.2 & 41.7 & 65.3 & 67.3 & 29.4 & 37.5 & 41.3 & 42.3 & 50.0 & 53.3 & & 49.2 & 53.1 \\
    \rowcolor{myhighlight} 
    \textbf{\vsa-8B} & & 61.7 & 59.7 & 68.0 & 67.1 & 39.3 & 33.4 & \textbf{69.6} & \textbf{69.9} & 34.1 & 36.2 & 46.2 & 43.0 & 56.5 & 55.8 & & 53.6 & 52.2\\
    \rowcolor{myhighlight} 
    \textbf{\vsa-I-8B} & &  \textbf{68.7} & \textbf{71.5} & \textbf{75.6} & \textbf{79.2} & 57.9 & 67.1 & 64.2 & 67.8 & \textbf{41.4} & \textbf{47.8} & \textbf{56.1} & \textbf{73.0} & \textbf{67.8} & \textbf{70.3} & & \textbf{61.7} & \textbf{68.1}\\
    
    \bottomrule
  \end{tabular}
  }
      \vspace{-0.15cm}
      \caption{Evaluation of selected models on \datasetlong for answer locating and post-hoc attribution. $\mathsf{F1}^\text{Q}_\text{box}$: answer locating F1 score; $\mathsf{F1}^\text{QA}_\text{box}$: post-hoc F1 score.}
  \label{tab:results2}
    \vspace{-0.4cm}
\end{table*}

\subsection{Visual Answer Grounding Benchmarking}

\tinytit{Zero-shot Models}
Table~\ref{tab:results} reports the zero-shot performance of sixteen representative MLLMs, evaluated on the seven public benchmarks constituting the test set of \datasetlong. Results are reported in terms of answer accuracy ($\mathsf{Acc}_\text{txt}$), Box F1 ($\mathsf{F1}_\text{box}$), and overall answer correctness ($\mathsf{Acc}$). Table~\ref{tab:results2} reports grounding F1 scores for answer-locating ($\mathsf{F1}^\text{Q}_\text{box}$) and post-hoc ($\mathsf{F1}^\text{QA}_\text{box}$) modes.

Overall, the largest models achieve the strongest results. However, $\mathsf{F1}_\text{box}$ scores remain consistently low, averaging below 40\% for $>$30B parameter models, below 20\% for 7-8B parameters models, and below 10\% for the smallest 2-3B parameters models. This underscores that even the best-performing models fail to reliably localize the visual evidence supporting their answers.

The complementary analysis in Table~\ref{tab:results2} reinforces these findings. When evaluated only on box generation, models achieve slightly higher localization precision than when required to jointly predict both the answer and its evidence location, indicating that the joint task is more challenging. The best-performing model, Qwen3-VL-32B, reaches nearly 47\% $\mathsf{F1}^\text{QA}_\text{box}$ in post-hoc mode on average across datasets, about 7 points higher than its average $\mathsf{F1}_\text{box}$ in the grounded answer generation setting, while maintaining stable answer accuracy. In answer-locating mode, most models underperform relative to post-hoc attribution, despite surpassing their $\mathsf{F1}_\text{box}$ from grounded answer generation.  This highlights the intrinsic difficulty of end-to-end grounded reasoning, which requires models to both reason and justify their outputs within a single forward pass.

\tit{Source Attribution Models}
Next, we evaluate models explicitly designed for grounding, including VISA-7B and our fine-tuned Qwen2.5-VL (\ours-7B), Qwen3-VL (\ours-8B), and InternVL3.5 (\ours-I-8B). As shown in Table~\ref{tab:results}, grounding supervision allows our models to consistently outperform both zero-shot MLLMs and VISA, achieving superior grounding ($\mathsf{F1}_\text{box}$) and overall metrics ($\mathsf{Acc}$). Notably, on the VISA test set, our models surpass the original VISA model in both grounding and overall performance. Despite operating at 7-8B parameters, our fine-tuned models rival or surpass significantly larger 32B and 38B models, demonstrating that grounding supervision is more impactful than scaling alone for attribution-intensive tasks. On average, overall answer correctness exceeds 30\%, more than 5 points above the best alternative without grounding supervision.

On the answer-locating ($\mathsf{F1}^\text{Q}_\text{box}$) and post-hoc attribution ($\mathsf{F1}^\text{QA}_\text{box}$) tasks (Table~\ref{tab:results2}), the fine-tuned models exhibit substantial gains over their zero-shot counterparts: on average, the Qwen2.5-based model improves by 31 points on both tasks, Qwen3-based model by 32 points on answer-locating and 23 points on post-hoc attribution, while the InternVL-based model gains 46 points on answer-locating and 49 points on post-hoc attribution.

Overall, \vsa-7B, \vsa-8B, and \vsa-I-8B achieve the highest scores across all evaluated datasets, surpassing even large general-purpose MLLMs. This trend holds across grounded answer generation, answer-locating, and post-hoc attribution, showing that \datasetlong supervision consistently improves grounding quality in MLLMs.

\section{Conclusion}
\label{sec:conclusion}
In this paper, we introduced \datasetlong (\dataset), an extensive benchmark with precise grounding annotations for evaluating visual answer grounding in Document VQA, and proposed \ours, a scalable Mask-based Perplexity-Derived Attribution framework for automatic, source-level attribution. Leveraging \ours, we unified multiple document understanding benchmarks to create a dataset that supports reliable evaluation and effective fine-tuning of grounding-capable multimodal models. Experiments showed that state-of-the-art MLLMs achieve strong textual accuracy but struggle to localize answer evidence; fine-tuning on \dataset substantially improves grounding across models and tasks. We expect \dataset to serve as a foundation for verifiable, interpretable document understanding and to foster models with explicit visual grounding.

\section*{Acknowledgments}
This work has been supported by the EU Horizon projects ``ELIAS'' (GA No. 101120237) and ``ELLIOT'' (GA No. 101214398), and by the EuroHPC JU project IT4LIA (No. 101234224). We also acknowledge the CINECA award under the ISCRA initiative, for the availability of high-performance computing resources and support.

\bibliography{egbib}
\end{document}


\maketitle

\begin{wrapfigure}{r}{0.48\textwidth}
    \centering
    \vspace{-0.4cm} 
    \begin{minipage}{0.95\linewidth}
        \centering

        \resizebox{\linewidth}{!}{
            \begin{tabular}{lcc}
                \toprule
                Split & Avg. Question Tokens & Avg. Answer Tokens \\
                \midrule
                Train & 12.34 & 18.36 \\
                Test  & 11.80 & 10.43 \\
                \bottomrule
            \end{tabular}
        }

        \vspace{0.2cm}

        \begingroup
            \raggedright
            \manualtablecaption{Average number of tokens per question and answer.}
            \label{tab:tokens}
            \par
        \endgroup

        \vspace{0.2cm}

        \begin{minipage}{0.49\linewidth}
            \centering
            \includegraphics[width=\linewidth]
                {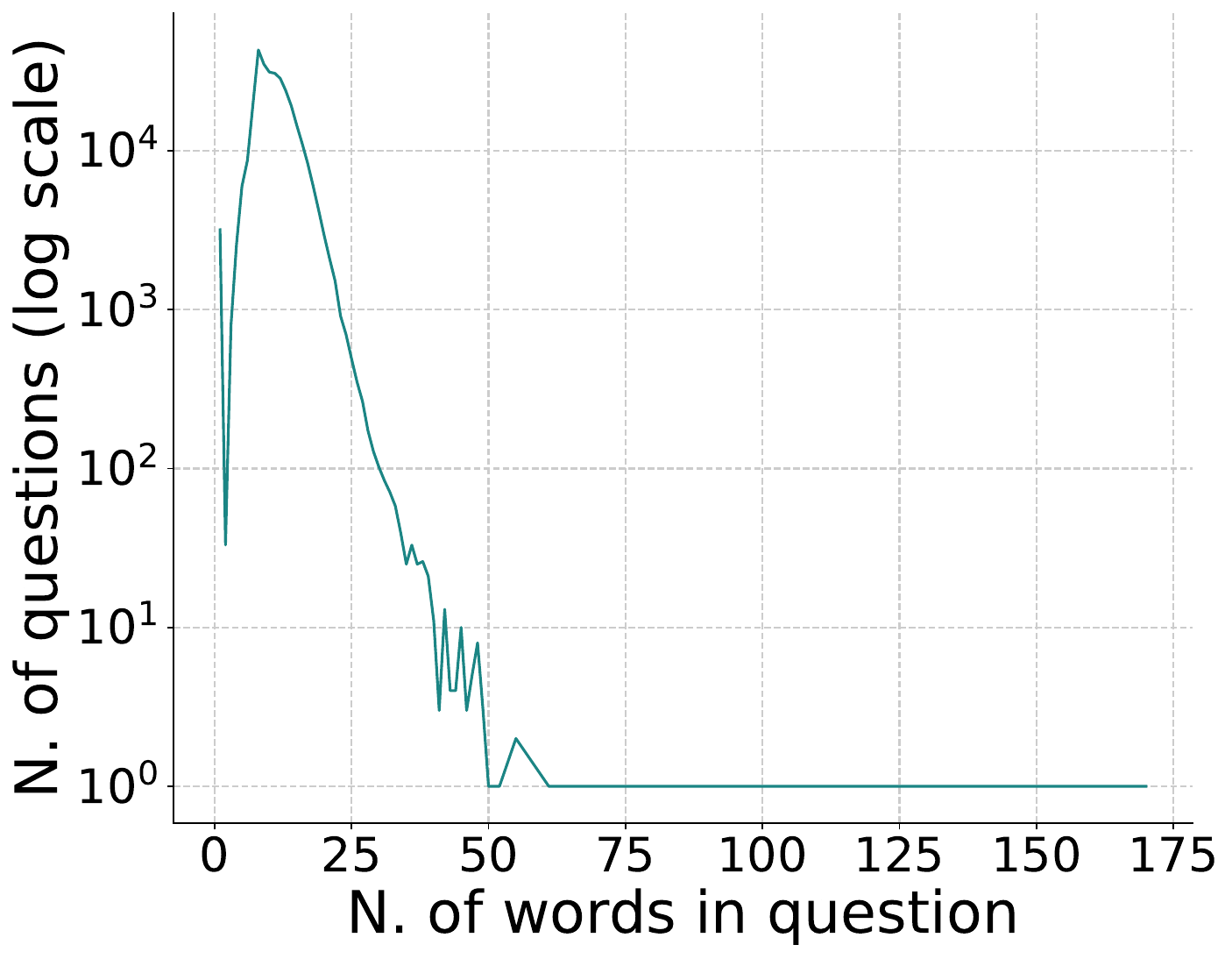}
        \end{minipage}
        \hfill
        \begin{minipage}{0.49\linewidth}
            \centering
            \includegraphics[width=\linewidth]
                {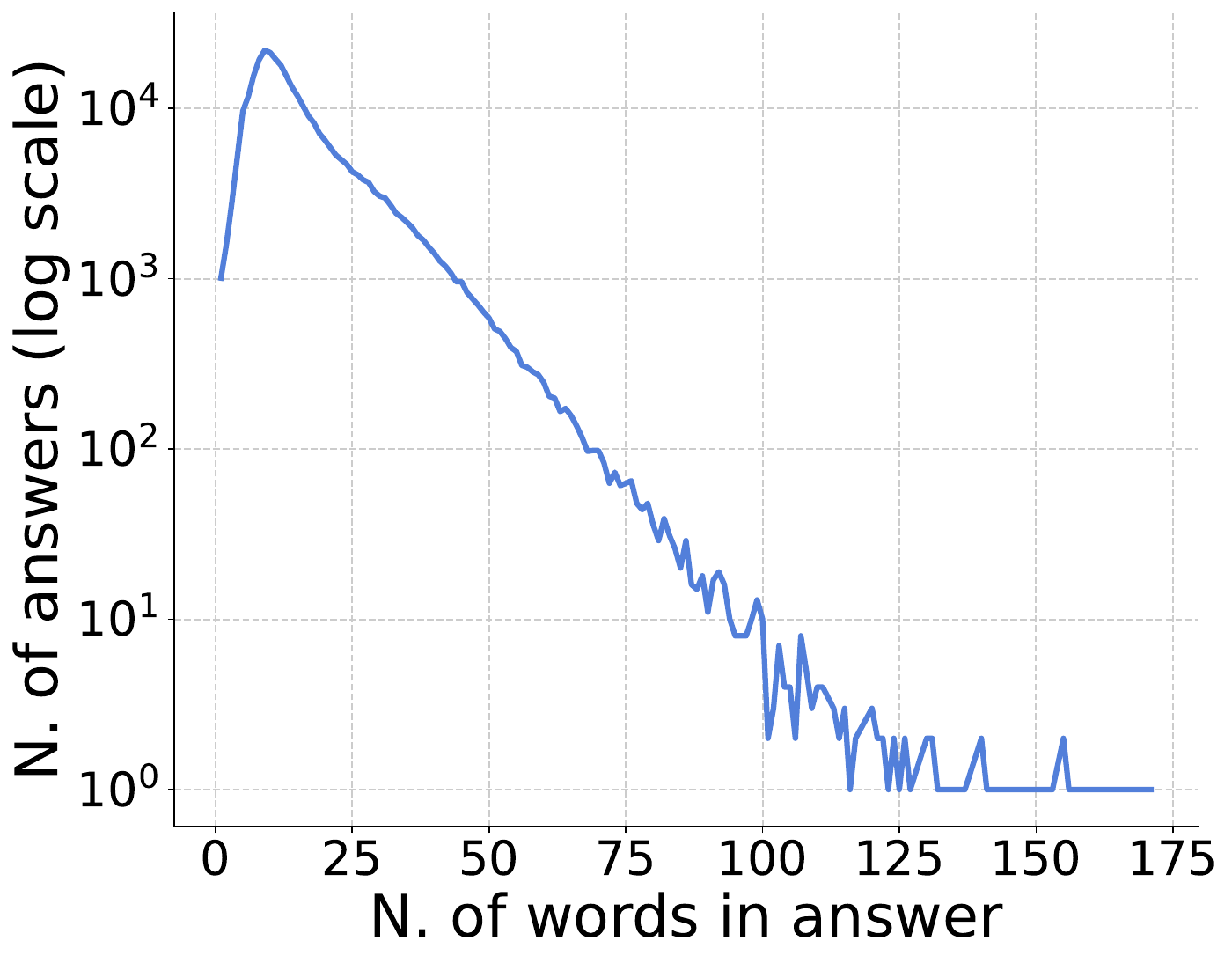}
        \end{minipage}

        \vspace{0.18cm}

        \caption{Word count distribution for questions (left) and answers (right).}
        \label{fig:qa_lengths}

    \end{minipage}

    \vspace{-0.5cm}
\end{wrapfigure}

\section{\datasetlong Details}

\datasetlong (\dataset) is our large-scale dataset for visual answer grounding, comprising 237,418 documents across diverse domains (scientific articles, business reports, presentation slides, and others), and 295,554 examples. Each sample consists of an image, a question-answer pair, a bounding box localizing the evidence region, and a region type label. The region types include \textit{paragraph/body}, \textit{caption}, \textit{heading/title}, \textit{subtitle/byline}, \textit{data}, \textit{sub-data}, \textit{table}, \textit{image}, \textit{picture}, \textit{list}, \textit{text}, and \textit{other}.

\begin{figure}[H]
    \centering
    \begin{minipage}[t]{0.48\linewidth}
        \centering
        \begin{minipage}{\linewidth}
            \centering

            \begin{minipage}{0.48\linewidth}
                \includegraphics[width=\linewidth]{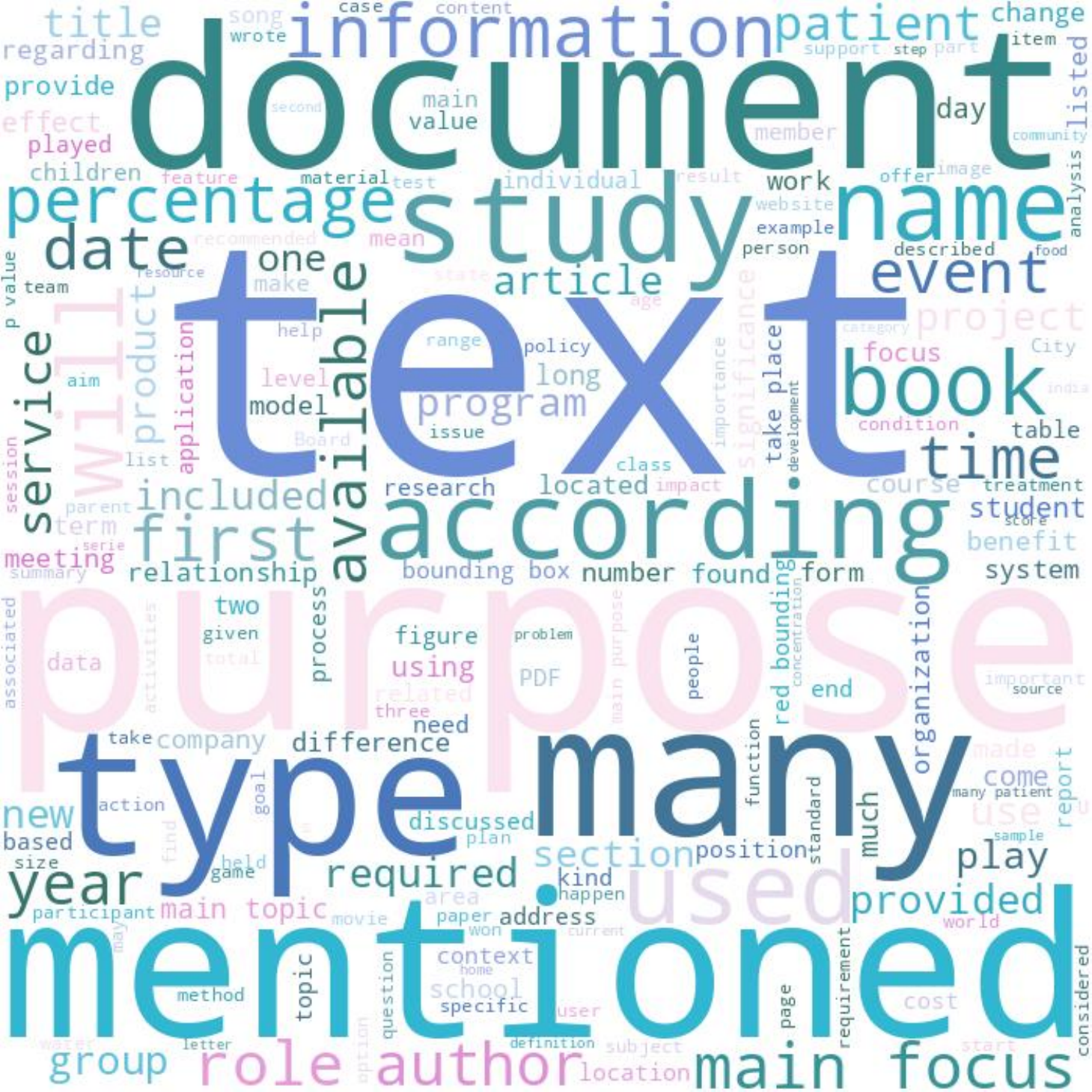}
            \end{minipage}
            \hfill
            \begin{minipage}{0.48\linewidth}
                \includegraphics[width=\linewidth]{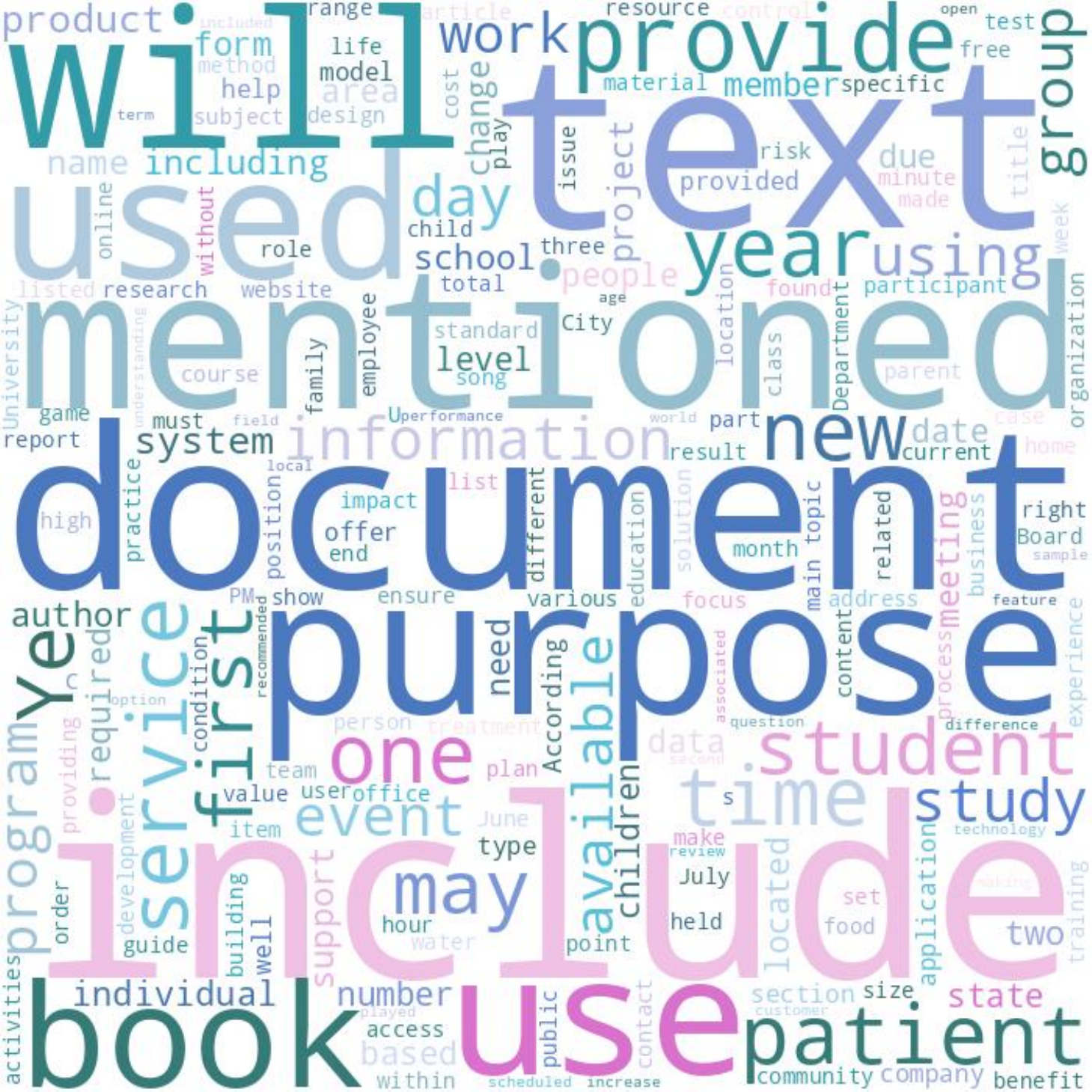}
            \end{minipage}

            \vspace{0.225cm}

            \caption{Word clouds for questions (left) and answers (right).}
            \label{fig:wordclouds}
        \end{minipage}
    \end{minipage}
    \hfill
    \begin{minipage}[t]{0.48\linewidth}
        \centering
        \begin{minipage}{\linewidth}
            \centering

            \begin{minipage}{0.49\linewidth}
                \centering
                \includegraphics[
                    width=\linewidth,
                    trim=8 5 6 5,
                    clip
                ]{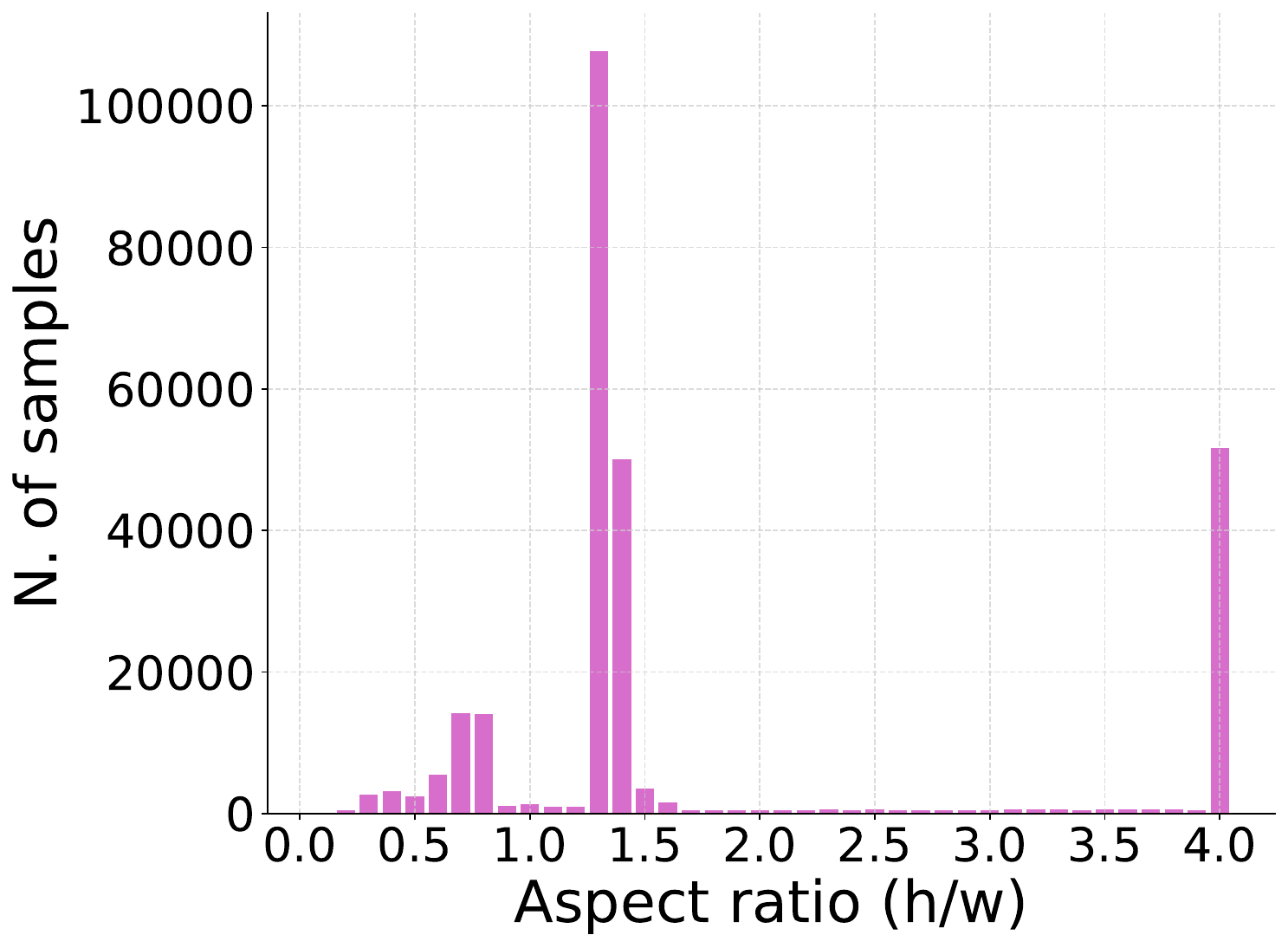}
            \end{minipage}
            \hfill
            \begin{minipage}{0.49\linewidth}
                \centering
                \includegraphics[
                    width=\linewidth,
                    trim=8 5 6 5,
                    clip
                ]{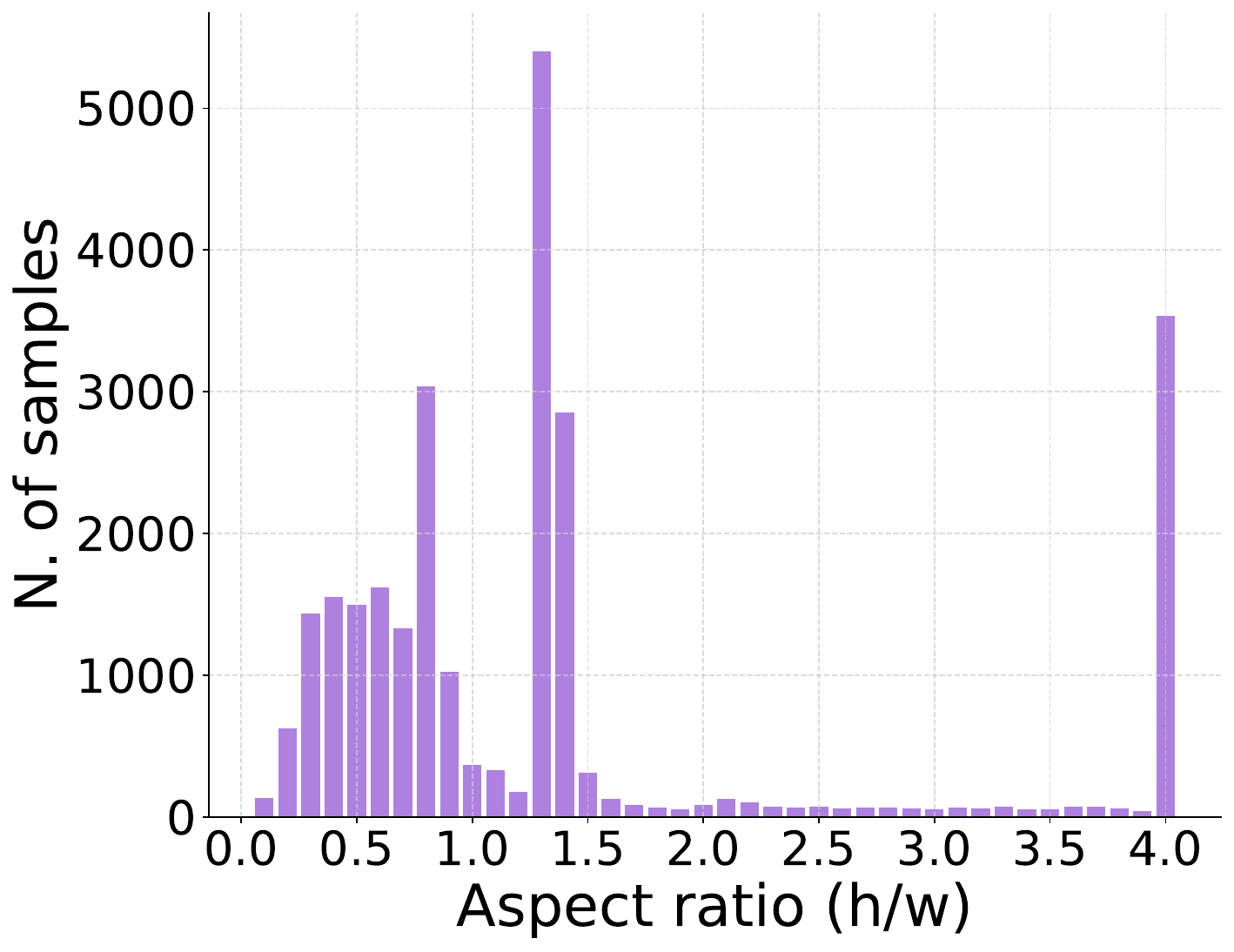}
            \end{minipage}

            \vspace{0.42cm}

            \caption{Distribution of images aspect ratios (height/width) for the train (left) and test (right) splits.}
            \label{fig:aspect_ratio}
        \end{minipage}
    \end{minipage}
\end{figure}

\begin{wrapfigure}{R}{0.5\linewidth}
    \centering
    \vspace{-0.47cm}
    \includegraphics[width=0.48\linewidth]{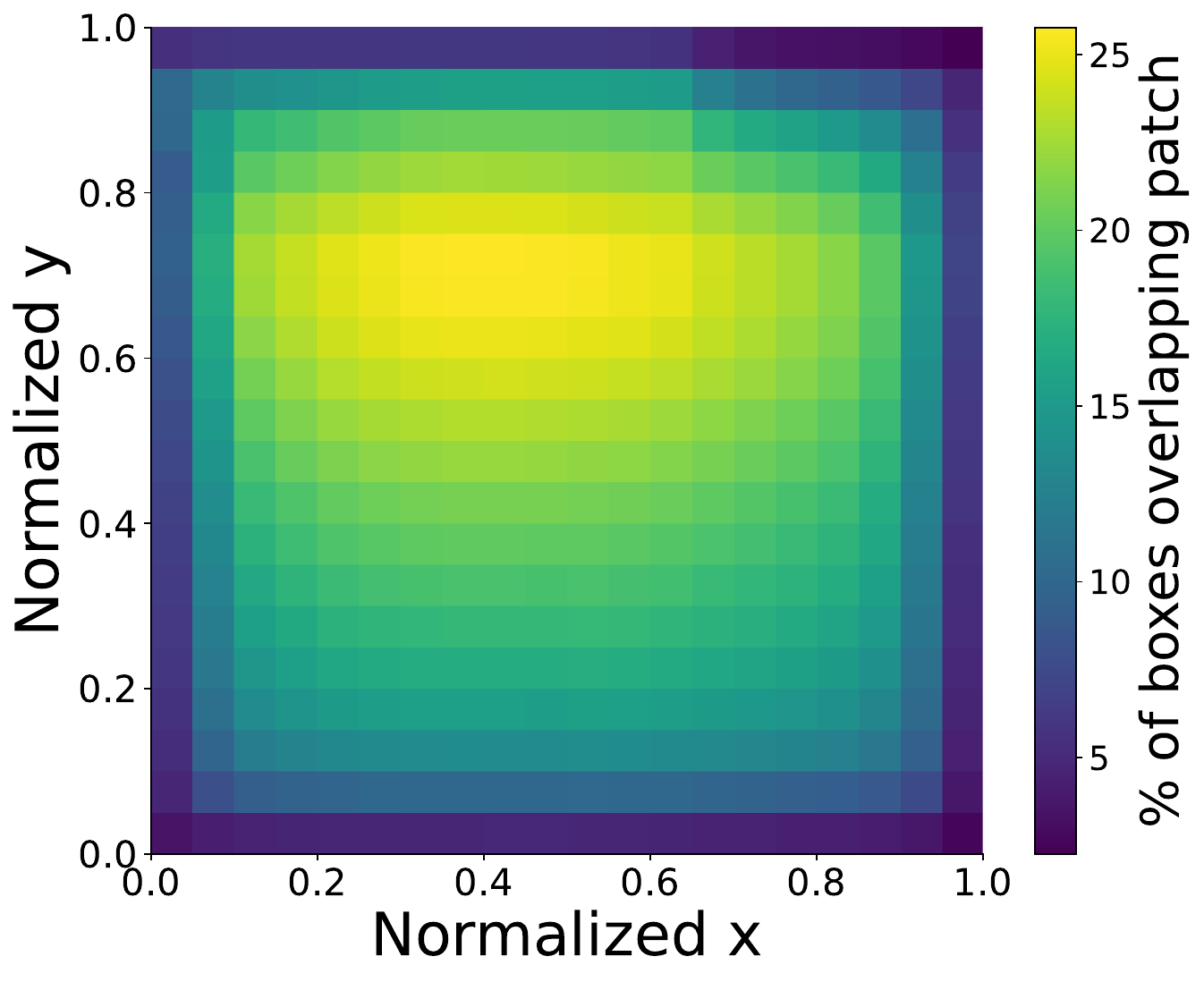}
    \hfill
    \includegraphics[width=0.48\linewidth]{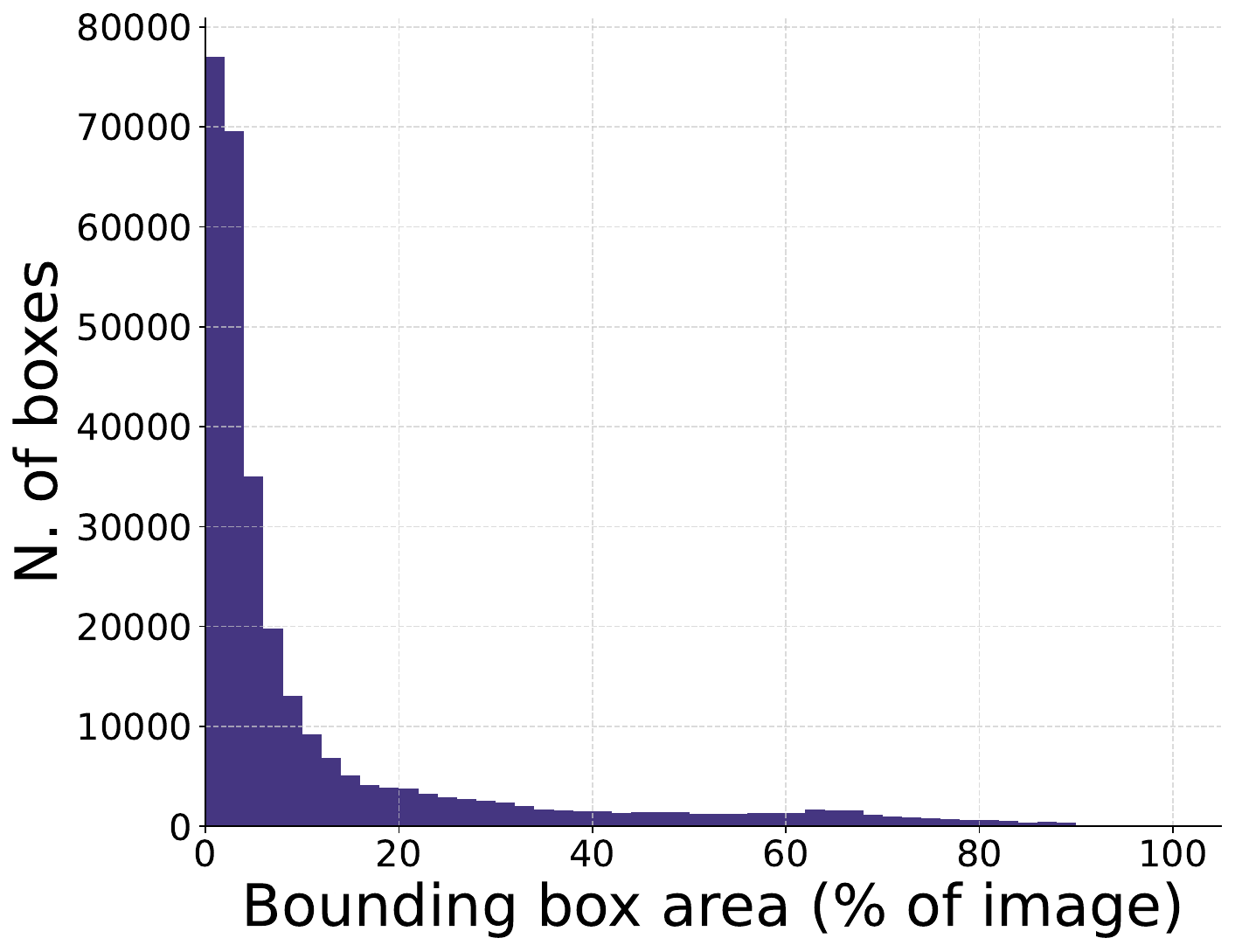}
    \vspace{-0.2cm}
    \caption{Heatmap of bounding-box coverage on a 20×20 grid (left) and histogram of bounding-box areas (right).}
    \label{fig:heatmap_hist}
    \vspace{-0.4cm}
\end{wrapfigure}

To further characterize the dataset, we provide additional visualizations. Figure~\ref{fig:qa_lengths} shows the distribution of question and answer lengths, and Table~\ref{tab:tokens} shows the corresponding average token counts. Figure~\ref{fig:wordclouds} presents word clouds with the most representative words from the dataset. 
The distribution of the image aspect ratios is shown in Figure~\ref{fig:aspect_ratio}, showing local modes corresponding to slides ($<1$), A4-like documents ($\sim1.5$), and rendered webpages ($>3$). We additionally examine the spatial distribution and scale of the annotated regions. First, each image is divided into a $20\times20$ grid of non-overlapping patches. Then, we count how many bounding boxes intersect with each patch. The resulting density map, along with the distribution of normalized region areas, is provided in Figure~\ref{fig:heatmap_hist}.
Annotated regions are primarily concentrated in the central-upper page areas, and most occupy only a small fraction of the page.

\subsection{Dataset Construction Prompts}
The answer abstraction step in \ours converts extractive answers from source datasets into an abstractive form using an MLLM. For the large-scale construction of \datasetlong, we use Qwen2.5-VL-7B-Instruct, selected for strong performance and efficient inference. For this model, as well as models evaluated in Table 3 of the main paper, we apply the prompt template shown in Figure~\ref{fig:abstractive_prompt_template} to generate abstractive answers.

For the attribution step, both during dataset construction and in the evaluation reported in Table 3, we adopt a minimal setup: the model receives the query as the prompt without system instructions and computes a score for the provided answer on the original image and its masked variants.

\subsection{Role of Answer Abstraction}

As discussed in the main paper, answer abstraction reformulates the predominantly extractive answers of the source datasets into a free-form style better aligned with current MLLMs, while providing sentence-level units for visual grounding.
Importantly, abstraction is not required by MAPPET itself: our ablation in Table 3 of the main paper shows comparable attribution accuracy with the original extractive answers, with a slight advantage for extractive answers before filtering. After filtering, however, abstractive answers consistently achieve higher attribution accuracy, suggesting that abstraction mainly improves the reliability of the retained annotations. Therefore, it should be viewed as a data harmonization and annotation-refinement step rather than as a necessary component of the MAPPET attribution mechanism.

\subsection{Dataset Licenses}
In the following, we report the licenses of the datasets from which \datasetlong is derived.

\begin{itemize}[topsep=2pt,itemsep=1pt]
    \item DoclingMatix~\cite{nassar2025smoldocling}: Community Data License Agreement Permissive 2.0
    \item DocVQA~\cite{Mathew_2021_WACV}: Apache License 2.0
    \item VisualMRC~\cite{tanaka2021visualmrc}: Creative Commons

    \item VISA~\cite{ma-etal-2025-visa}
    \begin{itemize}[topsep=2pt,itemsep=1pt]
        \item NQ: Apache License 2.0
        \item Wikipedia: Creative Commons Attribution Share Alike, GNU Free Documentation License family
        \item FineWeb-edu: Open Data Commons License Attribution family.
        \item PubLayNet: Community Data License Agreement - Permissive, Version 1.0.
    \end{itemize}

    \item SlideVQA~\cite{tanaka2023slidevqa}: The license permits the usage for testing and evaluation but does not allow distribution.

    \item VisualWebBench~\cite{liu2024visualwebbench}: Apache-2.0 license

    \item LongDocUrl~\cite{deng-etal-2025-longdocurl}: Apache-2.0 license

    \item MMLongBenchDoc~\cite{ma2024mmlongbench}: Apache-2.0 license
\end{itemize}

\subsection{Dataset Distribution}
Due to license restrictions, we will release the dataset excluding the SlideVQA portion, and the checkpoints fine-tuned on the dataset without it.

\subsection{\tasklong Examples}
\begin{figure*}[t]    
  \centering
    \centerline{\includegraphics[width=1\linewidth]{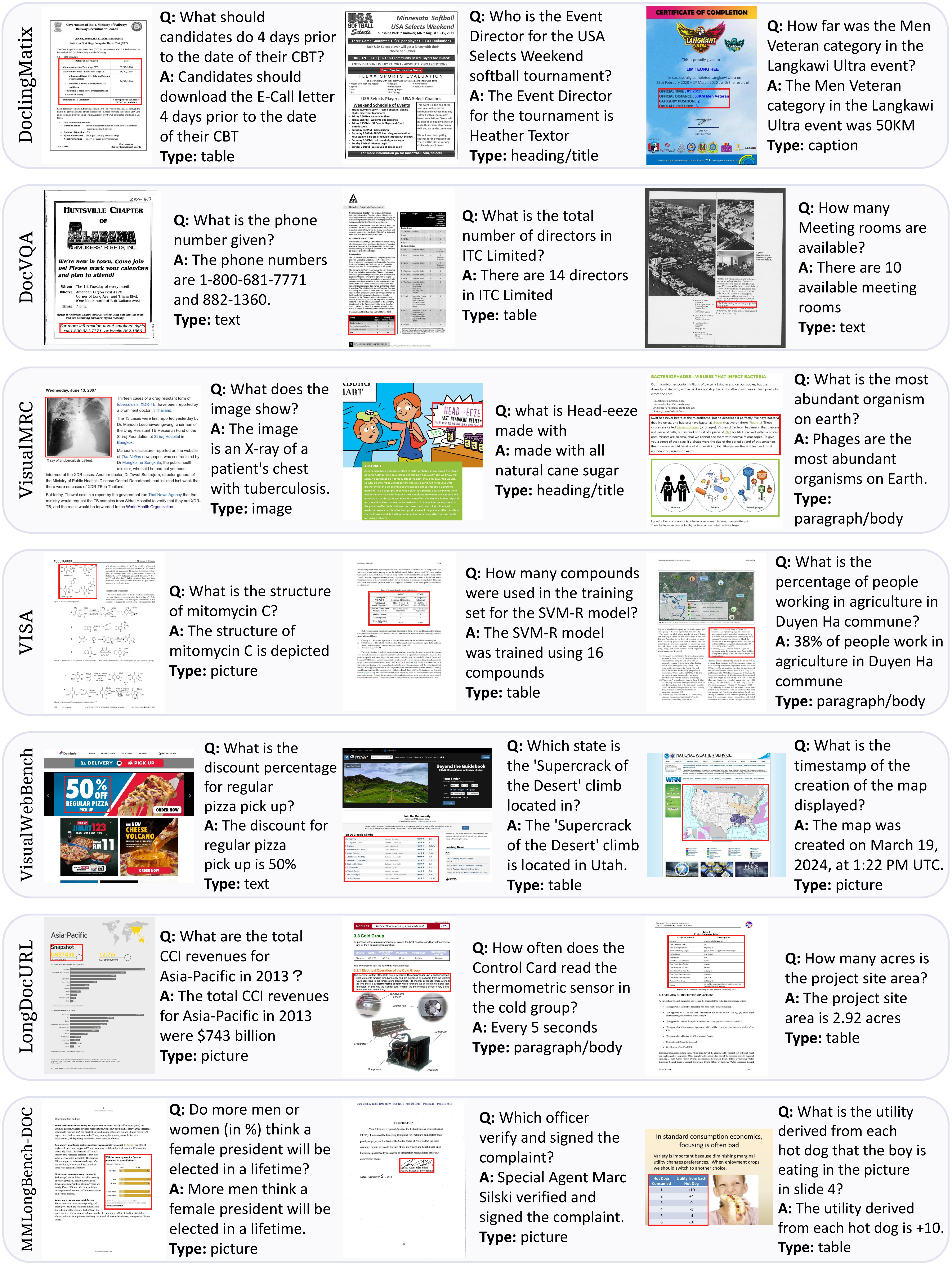}}
\vspace{-0.1cm}
  \caption{Qualitative examples for seven of the eight datasets that make up \datasetlong; SlideVQA is excluded due to licensing restrictions.}
  \label{fig:qualitatives}
\end{figure*}

We present qualitative examples of \datasetlong in Figure~\ref{fig:qualitatives} to highlight the quality of our attributions, the broad diversity of document types, and the variety of evidence regions associated with our annotation types. \dataset includes the following evidence types: \textit{caption}, which is text describing a figure or table; \textit{paragraph/body}, the main running text of the document; \textit{heading/title}, top-level structural titles or section headers; \textit{subtitle/byline}, secondary text under titles, including authors, affiliations, or dates; \textit{table}, any structured or unstructured tabular data with rows and columns; \textit{data}, standalone numeric or data-heavy content not clearly in a table; \textit{sub-data}, supporting or secondary numeric/text data tied to a main content element; \textit{list}, enumerated or bulleted lists; \textit{image}, generic visual elements like logos or decorative images; \textit{picture}, meaningful visual content such as photos, diagrams, charts, plots, or illustrations; \textit{text}, generic inline text that doesn’t fall into other text categories; and \textit{other}, any content not covered by the above categories (\eg, footers, headers, page numbers, or noise). 

Each example shows the document image alongside its question, answer, and annotation type, with the visual evidence highlighted by a red bounding box.

\section{Additional Implementation Details}
\label{supp:details}

\subsection{Evaluated Models}
The abstractive answers used for the experiments in Table~3 of the main paper are generated from the original extractive responses of the source datasets using vLLM, an efficient inference engine based on Paged Attention~\cite{kwon2023efficient}. The dataset abstractive answers are generated with the quantized version of Qwen2.5-VL-7B-Instruct\footnote{\href{https://huggingface.co/RedHatAI/Qwen2.5-7B-Instruct-quantized.w8a8}{\texttt{RedHatAI/Qwen2.5-7B-Instruct-quantized.w8a8}}}. The vLLM setup includes tensor parallelism with size 1, a maximum context length of 8,000 tokens, prefix caching for faster generation, a temperature of 0.1, and early stopping to prevent excessively long outputs.
\sloppy
For Qwen2.5-VL-32B, we employ the Qwen2.5-VL-32B-Instruct-AWQ model\footnote{\href{https://huggingface.co/Qwen/Qwen2.5-VL-32B-Instruct-AWQ}{\texttt{Qwen/Qwen2.5-VL-32B-Instruct-AWQ}}}, under vLLM with the same generation settings and a reduced concurrency configuration (\texttt{max\_num\_seqs = 2}, \texttt{max\_num\_batched\_tokens = 256}), together with \texttt{gpu\_memory\_utilization = 0.8} to accommodate the model.

For DeepSeek-VL-7B, we perform abstraction directly using the DeepSeek-VL-7B-Chat model\footnote{\href{https://huggingface.co/deepseek-ai/deepseek-vl-7b-chat}{\texttt{deepseek-ai/deepseek-vl-7b-chat}}} without additional quantization or modifications. Similarly, for Qwen3-VL-8B, we use the Qwen3-VL-8B-Instruct model\footnote{\href{https://huggingface.co/Qwen/Qwen3-VL-8B-Instruct}{\texttt{Qwen/Qwen3-VL-8B-Instruct}}} directly for generating the abstractive answers.
For computing the attribution, we use the standard models without additional modifications.

\subsection{Abstraction Statistics}
To assess the models ability to generate correct abstractive answers from the extractive counterpart, we sample 200 examples from the LongDocURL and DocVQA datasets and perform manual annotation. Three annotators are first instructed to identify semantic inconsistencies between the extractive and abstractive answers. Then, the annotators collectively re-evaluate and correct their annotations to reach unanimous agreement. In total, 4.5\% of the generated answers are hallucinated, consisting mostly of slight titles and names modifications. Overall, only 1.5\% of the generated answers are completely semantically dissimilar from the extractive counterparts.

\subsection{Zero-shot Prompt Templates}
Inference prompts are designed to generate answers and grounding coordinates from the selected models. Specifically, each prompt is designed to accomodate the original training configuration of each model. \textit{Grounded answer generation} prompts are designed to produce answers and to associate a bounding box with each generated sentence. Figure~\ref{fig:qwenvl_answer_box}, Figure~\ref{fig:internvl_answer_box}, and Figure~\ref{fig:visa_answer_box} show the template employed for the Qwen family, the InternVL family, and the VISA model, respectively. As far as \textit{post-hoc grounding} is concerned, prompt templates are designed to produce the coordinates that ground sentences to the document page and are shown in Figure~\ref{fig:qwen_box_from_asnwer}, Figure~\ref{fig:intervl_box_from_answer}, and Figure~\ref{fig:visa_box_from_answer}. Finally, prompt templates for \textit{answer-locating} are designed to answer the user query only with the required evidence. These templates are reported in Figure~\ref{fig:qwenvl_box}, Figure~\ref{fig:internvl_box}, and Figure~\ref{fig:visa_box}.

\subsection{Fine-tuning Prompt Templates}
\sloppy
The prompt templates used for fine-tuning Qwen2.5-VL, Qwen3-VL, and InternVL3.5-8B\footnote{\href{https://huggingface.co/OpenGVLab/InternVL3_5-8B-Instruct}{\texttt{OpenGVLab/InternVL3-5-8B-Instruct}}} are derived from those used for their zero-shot evaluation. Instructions about bounding box dimensions and rules reinforcing element-specific behaviors (\textit{e.g.}, how to generate boxes depending on the element type) are removed, assuming the models would learn them from the data. The resulting templates are used for both training and inference, excluding the assistant response during inference. The prompt templates for grounded answer generation, post-hoc grounding, and answer-locating are shown in Figure~\ref{fig:mappet_answer_box}, Figure~\ref{fig:mappet_box_from_answer}, and Figure~\ref{fig:mappet_box}, respectively.

\subsection{Resolution of Training Images}
During model fine-tuning, we randomize image resolutions with the following procedure. We first assign a target size for the longest side: 1024 pixels for images whose aspect ratio is at most 3:1, and 2048 pixels otherwise. We then add a random offset between 0 and 500 to the base value. The image is resized so that its longest side matches the resulting target length, while preserving the aspect ratio. After this step, we apply each model internal resizing logic.

\section{Box F1, Precision, and Recall Analysis}

The grounding metrics reported in Table 5 and Table 6 of the main paper use Box F1 as the primary measure of grounding accuracy. This choice is motivated by the generative nature of the evaluated MLLMs: although the DocAttriBench ground truth typically associates each answer with a single annotated evidence region, models are not constrained to predict exactly one bounding box and may instead output multiple candidate regions. In this setting, a single-box accuracy measure would not adequately account for over-prediction, as a model could identify the correct evidence while also producing several irrelevant regions. Box F1 explicitly captures both types of localization errors: after IoU-based matching, unmatched predictions are counted as false positives, while unmatched ground-truth regions are counted as false negatives. It therefore measures not only whether the supporting evidence is retrieved, but also whether it is localized without introducing spurious predictions.

To further characterize the grounding results and disentangle these two sources of error, we complement Box F1 with precision and recall for each task in Table~\ref{tab:grounded_answer_scores}, Table~\ref{tab:post_hoc_scores}, and Table~\ref{tab:answer_locating_scores}. In particular, this analysis allows us to determine whether the low zero-shot F1 scores observed for some models are mainly caused by a tendency to over-predict multiple boxes or by a failure to localize the correct supporting evidence.

The numerical results reveal a consistent pattern for Qwen3-VL-2B: it frequently generates multiple boxes for the grounded answer generation task, inflating false positives and leading to low precision. Qualitative inspection shows that these extra boxes rarely correspond to meaningful reasoning, instead, the model produces several regions without an actual answer. This behavior largely explains the poor performance observed for this model.

In contrast, we do not observe substantial drops in either precision or recall for the other models. Hence, the observed low F1 scores primarily reflect challenges in locating the correct evidence, rather than systematic over-generation or a breakdown of the reasoning process.

Across tasks, the fine-tuned models consistently outperform their zero-shot counterparts. In grounded answer generation, \ours-8B improves over the best zero-shot model (Qwen3-VL-32B) by 26 points in average precision and 23 points in average recall, and by roughly 50 points in both metrics compared to its backbone (Qwen3-VL-8B). In the post-hoc attribution task, \ours-I-8B exceeds the best zero-shot model by about 22 points in both metrics and improves over its backbone (InternVL-3.5-8B) by over 50 points. Finally, in answer locating, \ours-I-8B surpasses Qwen3-VL-32B by approximately 19 points in both precision and recall, and its backbone by more than 46 points.

Overall, the fine-tuned models not only boost F1 but do so through meaningful gains in both precision and recall, highlighting a markedly stronger capability to locate the correct supporting regions for their answers.

\section{Qualitative Results}
\label{supp:qualitatives}

\begin{figure*}[t]    
  \centering
    \centerline{\includegraphics[width=0.99\linewidth]{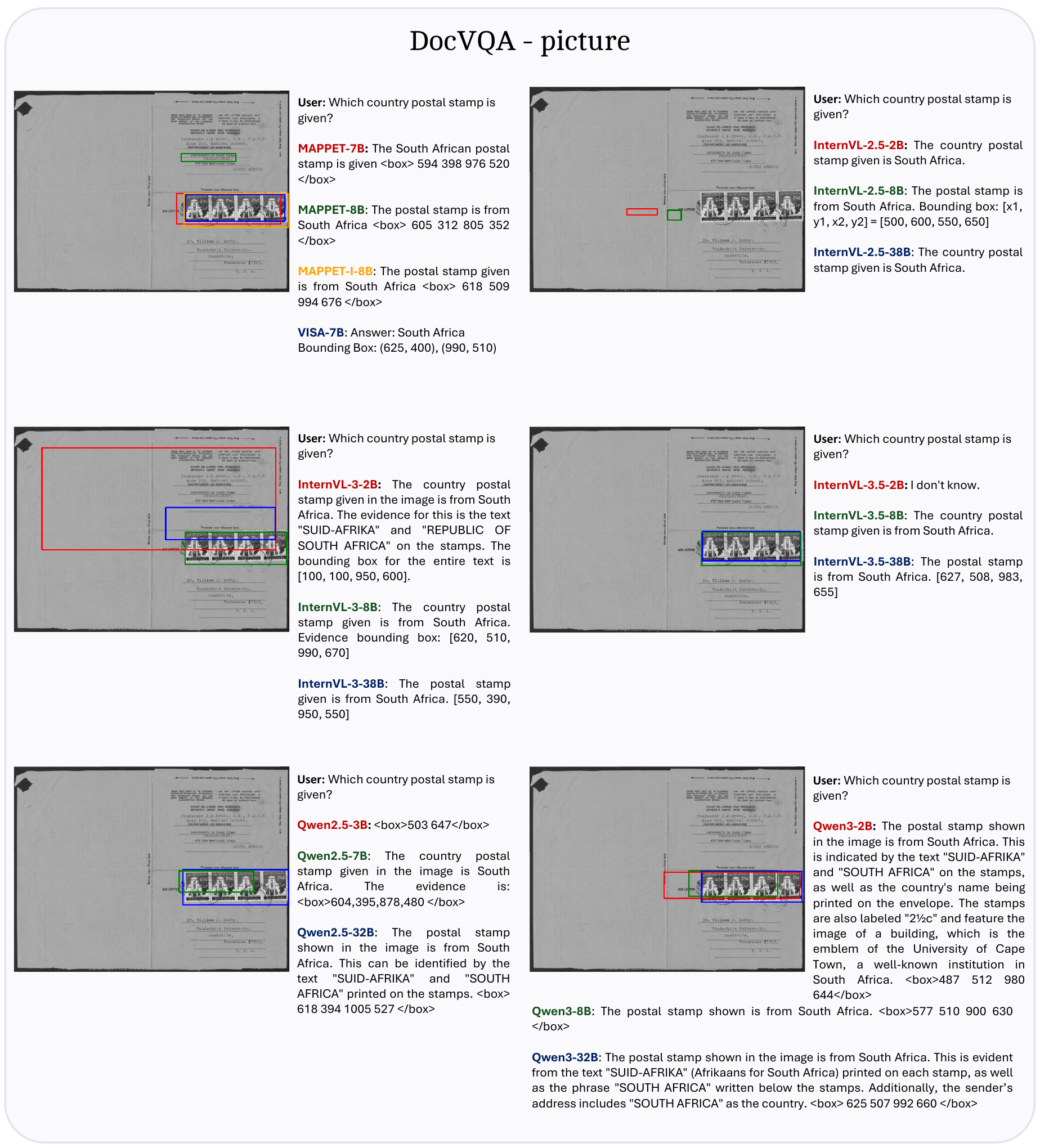}}
\vspace{-0.1cm}
  \caption{Grounded answer generation with multiple models on the DocVQA dataset.}
  \label{fig:model_examples_docvqa_picture}
\end{figure*}

\begin{figure*}[t]    
  \centering
    \centerline{\includegraphics[width=0.99\linewidth]{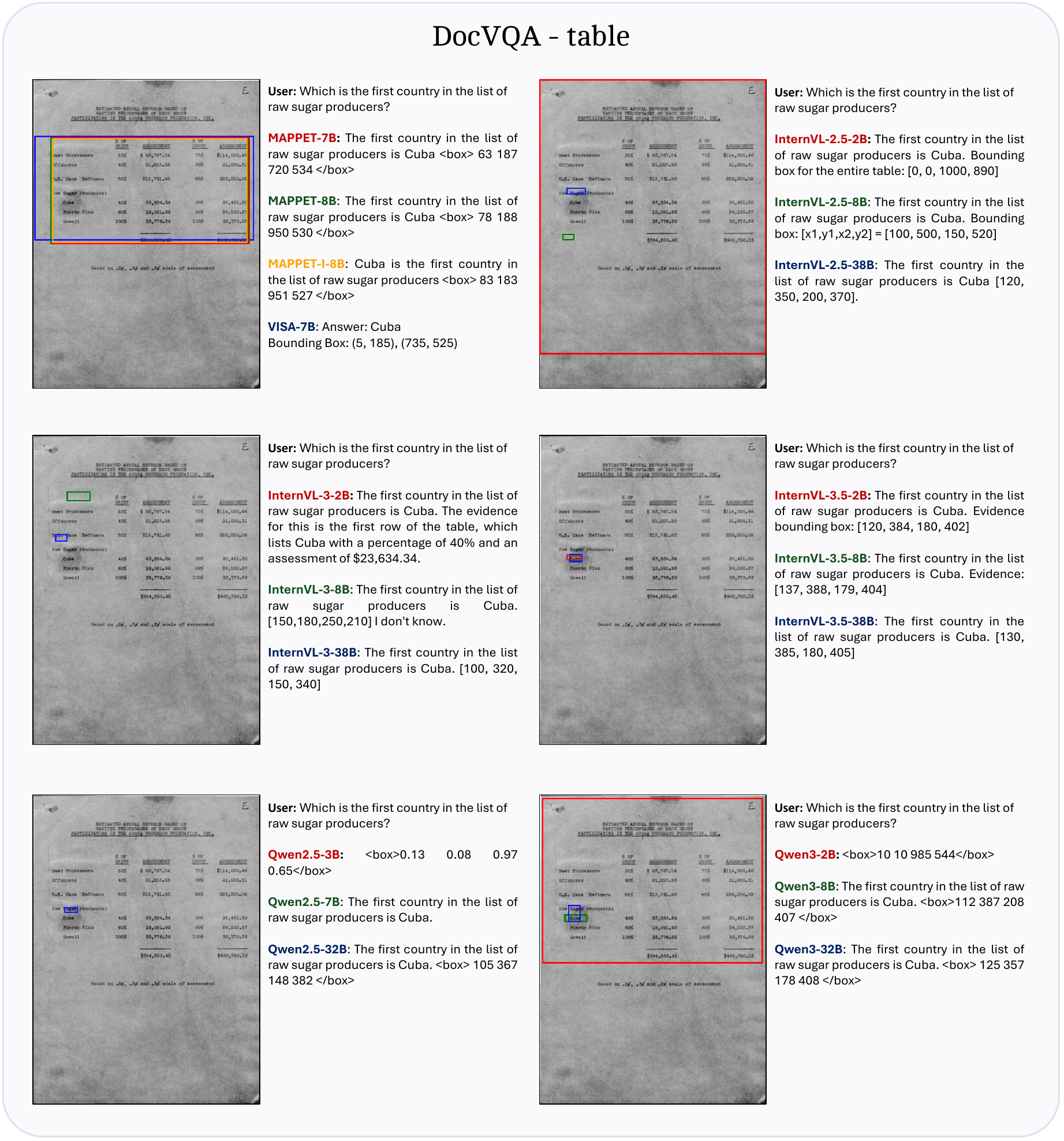}}
\vspace{-0.1cm}
  \caption{Grounded answer generation with multiple models on the DocVQA dataset.}
  \label{fig:model_examples_docvqa_table}
\end{figure*}

\begin{figure*}[t]    
  \centering
    \centerline{\includegraphics[width=0.99\linewidth]{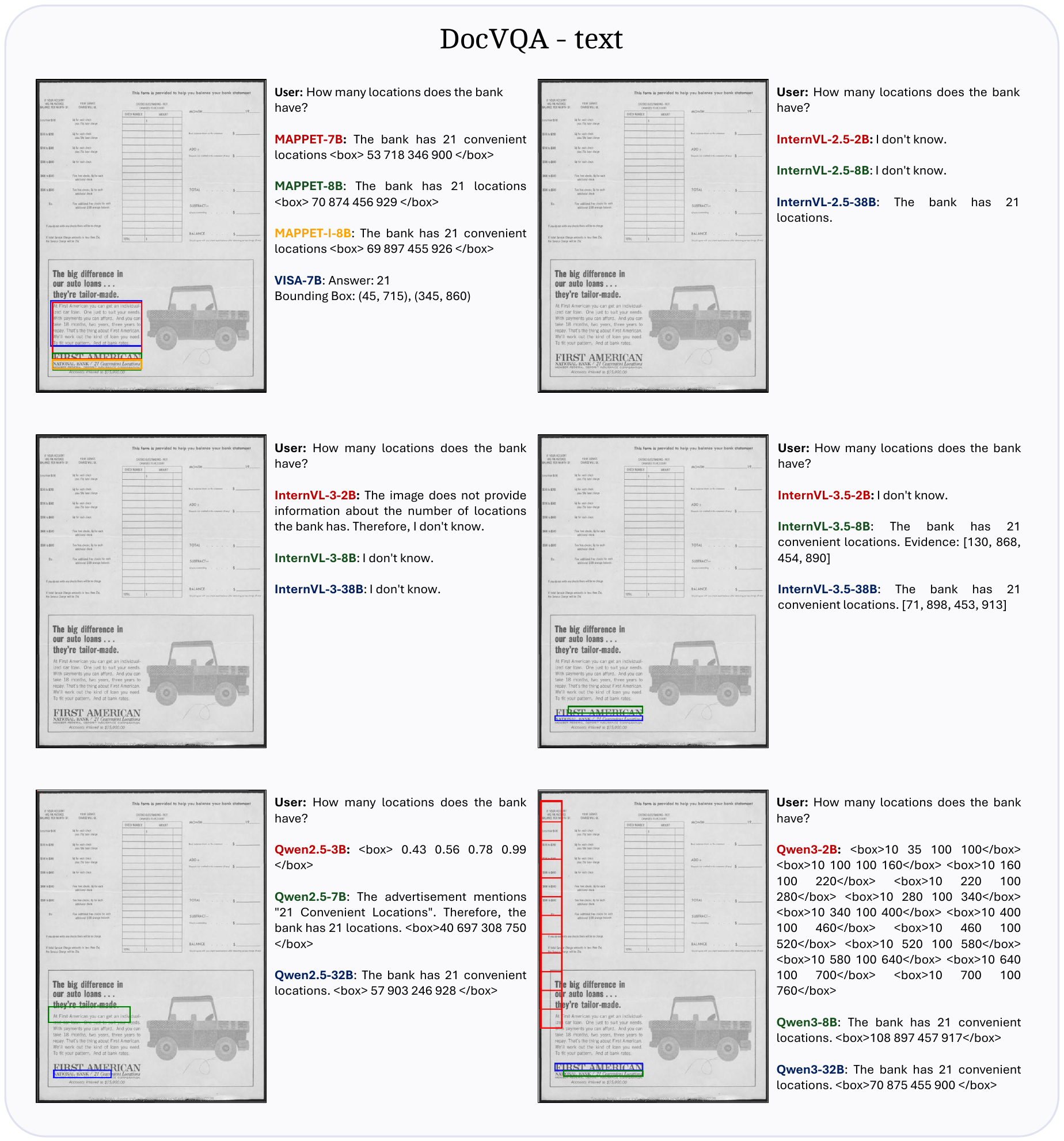}}
\vspace{-0.1cm}
  \caption{Grounded answer generation with multiple models on the DocVQA dataset.}
  \label{fig:model_examples_docvqa_text}
\end{figure*}

\begin{figure*}[t]    
  \centering
    \centerline{\includegraphics[width=0.99\linewidth]{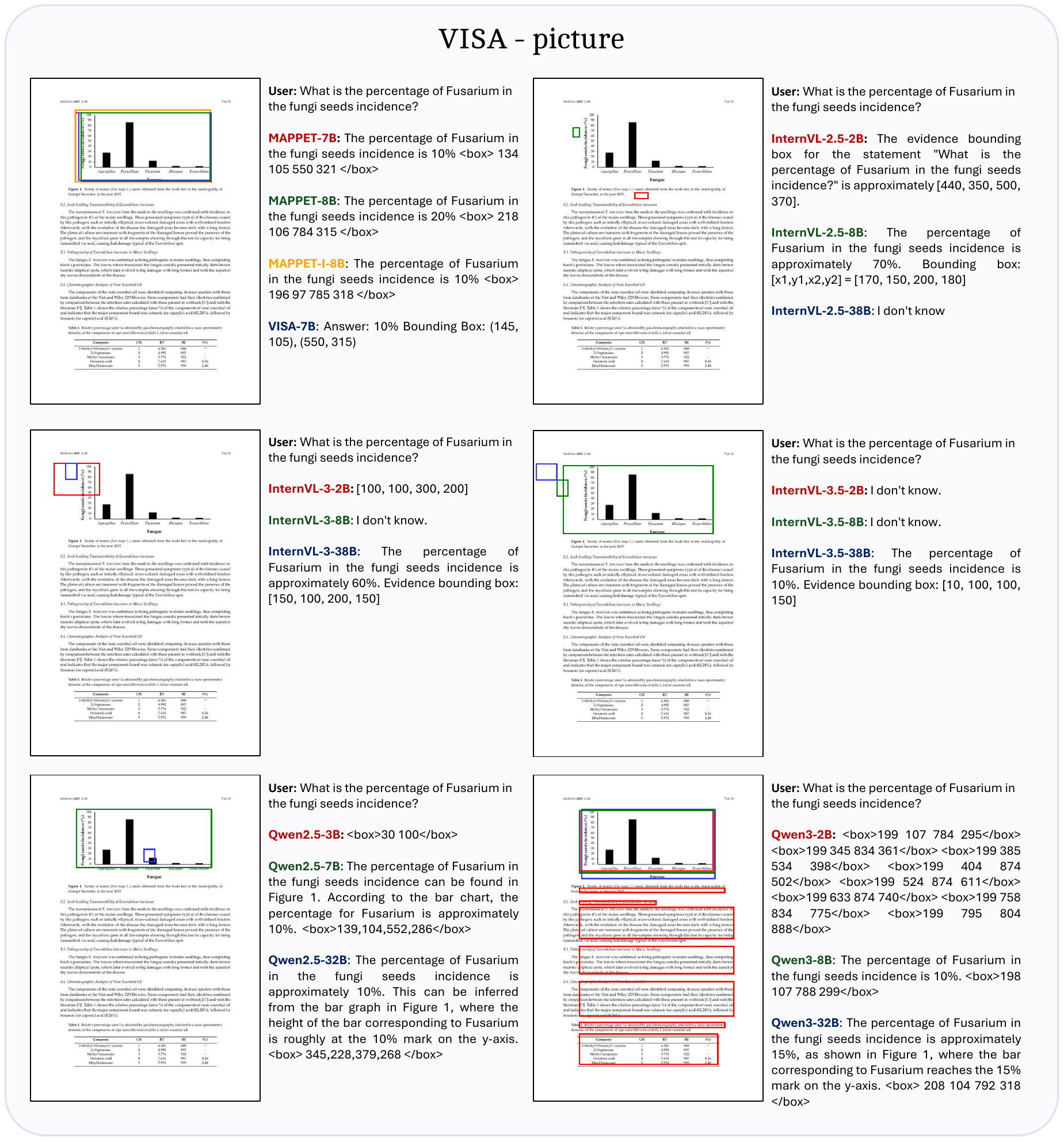}}
\vspace{-0.1cm}
  \caption{Grounded answer generation with multiple models on the VISA dataset.}
  \label{fig:model_examples_visa_picture}
\end{figure*}

\begin{figure*}[t]    
  \centering
    \centerline{\includegraphics[width=0.99\linewidth]{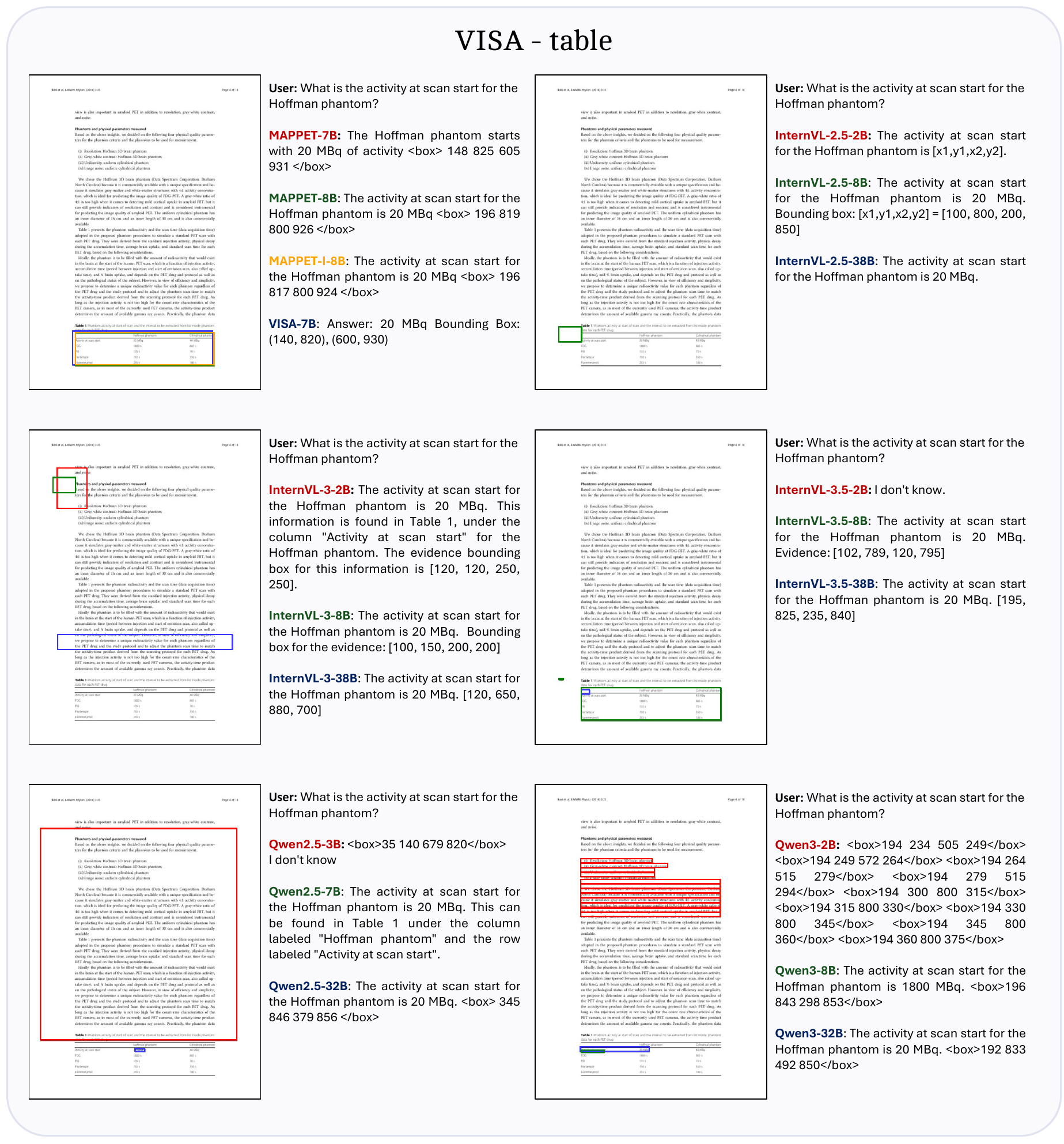}}
\vspace{-0.1cm}
  \caption{Grounded answer generation with multiple models on the VISA dataset.}
  \label{fig:model_examples_visa_table}
\end{figure*}

\begin{figure*}[t]    
  \centering
    \centerline{\includegraphics[width=0.99\linewidth]{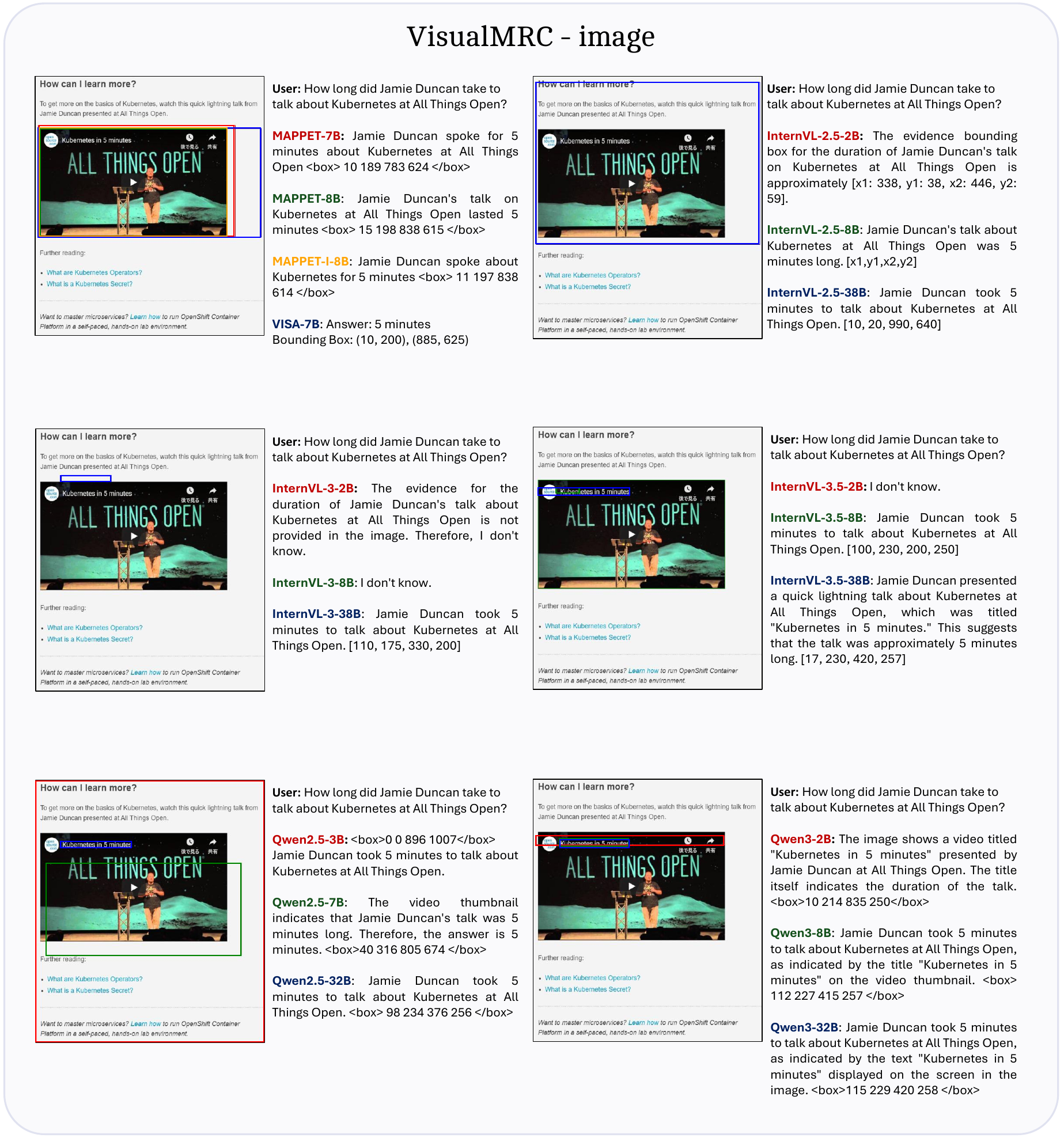}}
\vspace{-0.1cm}
  \caption{Grounded answer generation with multiple models on the VisualMRC dataset.}
  \label{fig:model_examples_visualmrc_picture}
\end{figure*}

\begin{figure*}[t]    
  \centering
    \centerline{\includegraphics[width=0.99\linewidth]{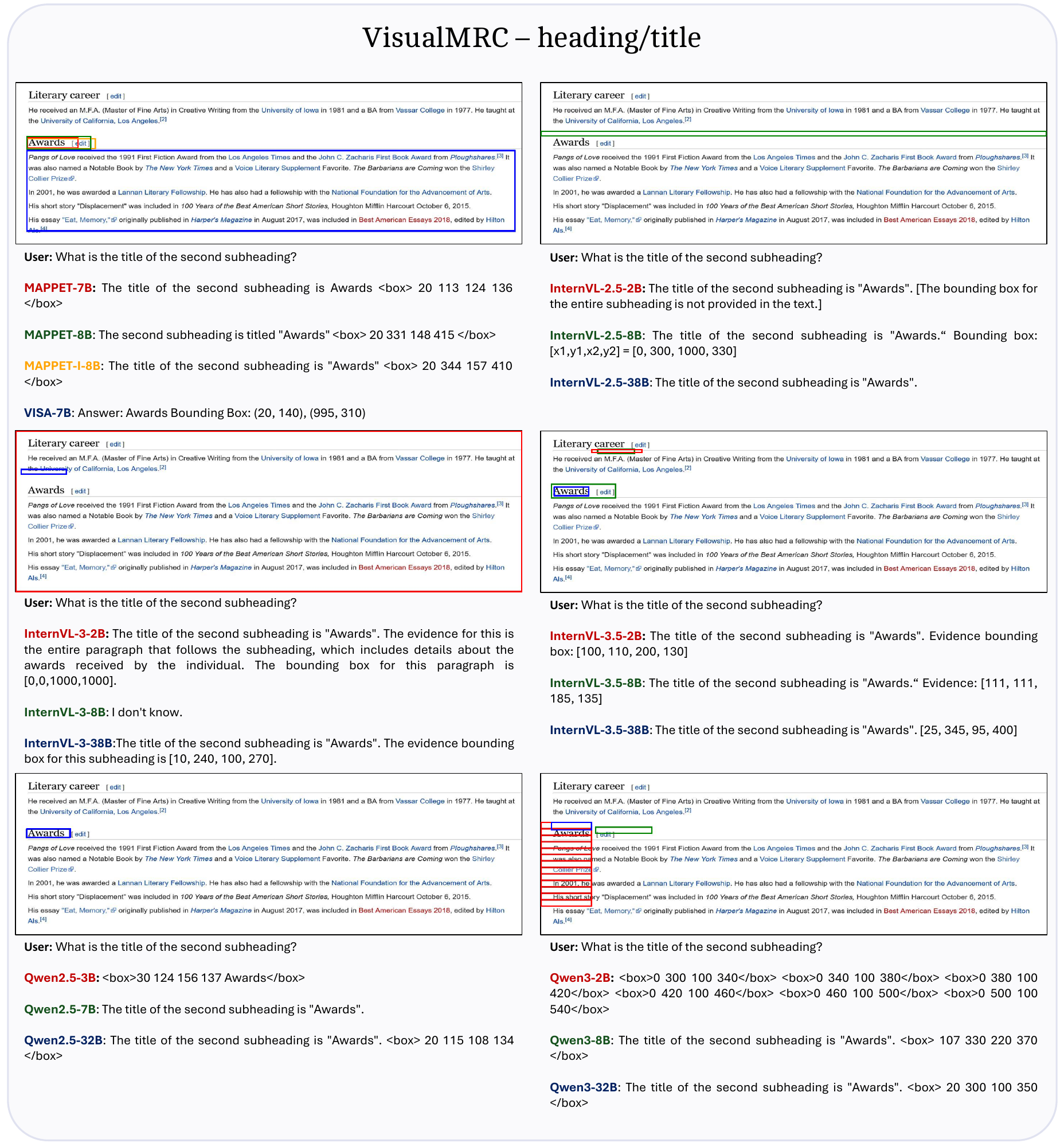}}
\vspace{-0.1cm}
  \caption{Grounded answer generation with multiple models on the VisualMRC dataset.}
  \label{fig:model_examples_visualmrc_text}
\end{figure*}

We provide qualitative results to illustrate how models fine-tuned on our dataset behave across different document types, compared to VISA and zero-shot models. From Figure~\ref{fig:model_examples_docvqa_picture} to Figure~\ref{fig:model_examples_visualmrc_text} we present examples spanning a range of evidence types. For a subset of the source datasets and evidence types (\textit{e.g.}, DocVQA-text), we display the input image alongside the model-generated answer and its visual attribution. The evidence region predicted by each model is highlighted in the color corresponding to that model.
The outputs of the explicitly grounded models, namely our fine-tuned models \ours-7B, \ours-8B, \ours-I-8B, and VISA-7B, are shown together at the top left of the images, demonstrating the high precision achieved by our models, often surpassing that of VISA.
Overall, fine-tuning on our dataset consistently improves visual grounding quality, reduces hallucinated attributions, and demonstrates high localization precision.

\section{Paper-VISA Error Analysis}
Paper-VISA~\cite{ma-etal-2025-visa} is among the few approaches for automatic \task dataset construction, and is the only one relying on open-source components. The original system follows a three-step pipeline. First, an off-the-shelf document layout analysis model identifies structural elements from the document page. Second, one element is sampled and highlighted on the document image with a red bounding box. Finally, an MLLM is tasked with the generation of a question, that can be answered using the content of the bounding box, together with the corresponding answer. To assess the reliability of this procedure, we manually inspected 100 single-page examples sampled from the Paper-VISA dataset. The detected error rates are reported in Table~\ref{tab:papervisa_errors}. 

\begin{wraptable}{r}{0.3\textwidth}
\centering
\vspace{-0.55cm}
\resizebox{0.95\linewidth}{!}{
\begin{tabular}{lr}
\toprule
Error type & Quantity \\
\midrule
Hallucination & 10\% \\
Red Box & 4\% \\
Wrong Box & 23\% \\
Overall & 33\% \\
\bottomrule
\end{tabular}
}
\vspace{0.15cm}
\caption{Error rates for Paper-VISA.}
\label{tab:papervisa_errors}
\vspace{-0.4cm}
\end{wraptable}

Among the inspected examples, 10\% contain hallucinated answers (\textit{Hallucination}), either referring to content not present in the page or reporting incorrect or incomplete numbers, figures, or names; 4\% of the questions refer to the red bounding box overlayed on top of the image during the dataset generation process (\textit{Red Box}), which is unavailable at inference time and creates unsolvable examples; 23\% of the samples exhibit wrong grounding (\textit{Wrong Box}), \textit{i.e.}, the generated question-answer pair is not aligned with the bounding box that was overlayed to the document image. Finally, 33\% of the inspected samples contain at least one of the aforementioned errors. While the hallucinated examples can still be used for training and evaluating post-hoc grounding approaches, approximately 23\% of the dataset contains wrong answer-box associations, introducing noise that makes the approach less suitable for both model training and evaluation.

\section{Limitations and Impact}
\label{supp:limitations}

While MAPPET demonstrates promising results for the automatic construction of large-scale datasets for grounded answers generation, there remain opportunities for further improvement. First, although MAPPET can be used for the generation of multi-sentence datasets, our experiments are mostly limited to single-sentence answers and single-page documents. Second, we assume that each answer can be localized within a single region of the document image, which limits the dataset effectiveness in cases where evidence must be gathered from multiple regions.

On the other hand, MAPPET enables the zero-shot creation of large-scale datasets with limited computational resources. Most importantly, by producing domain-specific training data, it allows existing document-question-answering models to be fine-tuned efficiently and adapted quickly to new domains.

\begin{figure*}[t]
    \centering
    \begin{tcolorbox}[
        colback=promptbg,
        colframe=promptborder,
        arc=2mm,
        boxrule=1pt,
        width=\linewidth,
        left=5pt, right=5pt, top=5pt, bottom=5pt,
         fontupper=\ttfamily\scriptsize
    ]
 
{\sffamily\bfseries System:} You are an agent excellent at identifying evidence in a document page.\\
Your job is to answer the user's query based on the provided image and the context.\\
When you answer, provide evidence bounding boxes in the format <box> x1 y1 x2 y2 </box>.\\
Add an evidence bounding box at the end of each generated sentence.\\
\\
If you cannot find the answer, respond with "I don't know".\\

{\sffamily\bfseries User:} \{query\}\\

{\sffamily\bfseries Assistant:} \{answer\}
    \end{tcolorbox}
    \vspace{-10pt}
    \caption{Fine-tuning prompt for grounded answer generation.}
    \label{fig:mappet_answer_box}
    \vspace{-0.3cm}
\end{figure*}

\begin{figure*}[t]
    \centering
    \begin{tcolorbox}[
        colback=promptbg,
        colframe=promptborder,
        arc=2mm,
        boxrule=1pt,
        width=\linewidth,
        left=5pt, right=5pt, top=5pt, bottom=5pt,
        fontupper=\ttfamily\scriptsize
    ]
     
{\sffamily\bfseries System:} You are an agent excellent at identifying evidence in a document page.\\
Your job is to identify the answer contained in the user's input based on the provided image and the context.\\
Respond by providing the evidence bounding boxes in the format <box> x1 y1 x2 y2 </box>.\\
The input is constituted by the user query and the answer to that query.\\
\\
If you cannot find the answer, respond with the empty box <box> </box>.\\

{\sffamily\bfseries User:} Query:\\
\{query\}\\
\\
Answer:\\
\{answer\}\\

{\sffamily\bfseries Assistant:} \{bounding\_box\}
    \end{tcolorbox}
    \vspace{-10pt}
    \caption{Fine-tuning prompt for post-hoc grounding.}
    \label{fig:mappet_box_from_answer}
    \vspace{-0.4cm}
\end{figure*}

\begin{figure*}[t]
    \centering
    \begin{tcolorbox}[
        colback=promptbg,
        colframe=promptborder,
        arc=2mm,
        boxrule=1pt,
        width=\linewidth,
        left=5pt, right=5pt, top=5pt, bottom=5pt,
         fontupper=\ttfamily\scriptsize
    ]
    
{\sffamily\bfseries System:} You are an agent excellent at identifying evidence in a document page.\\
Your job is to locate the answer to the user's query based on the provided image and the context.\\
Respond by providing the evidence bounding boxes in the format <box> x1 y1 x2 y2 </box>.\\
\\
If you cannot find the answer, respond with the empty box <box> </box>.\\

{\sffamily\bfseries User:} \{query\}\\

{\sffamily\bfseries Assistant:} \{bounding\_box\}
    \end{tcolorbox}
    \vspace{-10pt}
    \caption{Fine-tuning prompt for answer-locating.}
    \label{fig:mappet_box}
    \vspace{-0.3cm}
\end{figure*}

\begin{figure*}[t]
    \centering
    \begin{tcolorbox}[
        colback=promptbg,
        colframe=promptborder,
        arc=2mm,
        boxrule=1pt,
        width=\linewidth,
        left=5pt, right=5pt, top=5pt, bottom=5pt,
         fontupper=\ttfamily\scriptsize
    ]
         
{\sffamily\bfseries User:} Consider the following query:\\
       \{query\}\\

        Here is the extractive answer to the query:\\
        \{answer\}\\

        Rephrase the answer in a short sentence. Include all numbers and names.\\
        Answer with a JSON of the following form:\\
        \{\{\\
        \hspace*{1em} "answer": "$<$The old answer to the query$>$",\\
        \hspace*{1em} "abstracted\_answer": "$<$The rephrased answer$>$"\\
        \}\}
    \end{tcolorbox}
    \vspace{-10pt}
    \caption{Prompt used for MLLMs during the automatic annotation pipeline, in the answer abstraction phase.}
    \label{fig:abstractive_prompt_template}
    \vspace{-0.3cm}
\end{figure*}

\begin{figure*}[t]
    \centering
    \begin{tcolorbox}[
        colback=promptbg, 
        colframe=promptborder,
        arc=2mm,
        boxrule=1pt,
        width=\linewidth,
        left=5pt, right=5pt, top=5pt, bottom=5pt,
        fontupper=\ttfamily\scriptsize
    ]
     
{\sffamily\bfseries System:} You are an agent excellent at identifying evidence in a document page.\\
Your job is to answer the user's query based on the provided image and the context.\\
When you answer, provide evidence bounding boxes in the format <box> x1 y1 x2 y2 </box> where (x1, y1) is the top-left corner.\\
Add an evidence bounding box at the end of each generated sentence.\\
\\
If you cannot find the answer, respond with "I don't know".\\
\\
If the evidence is part of a paragraph, return the bounding box of the entire paragraph.\\
If the evidence is part of a table, return the bounding box of the entire table.\\
If the evidence is part of a figure, return the bounding box of the entire figure.\\

{\sffamily\bfseries User:} \{query\}
    \end{tcolorbox}
    \vspace{-10pt}
    \caption{Grounded answer generation prompt used for the Qwen family.}
    \label{fig:qwenvl_answer_box}
    \vspace{-0.3cm}
\end{figure*}

\begin{figure*}[t]
    \centering
    \begin{tcolorbox}[
        colback=promptbg,
        colframe=promptborder,
        arc=2mm,
        boxrule=1pt,
        width=\linewidth,
         left=5pt, right=5pt, top=5pt, bottom=5pt,
         fontupper=\ttfamily\scriptsize
    ]
     
{\sffamily\bfseries User:} You are an agent excellent at identifying evidence in a document page.\\
Your job is to answer the user's query based on the provided image and the context.\\
When you answer, provide evidence bounding boxes in the format [x1,y1,x2,y2] where (x1, y1) is the top-left corner.\\
Add an evidence bounding box at the end of each generated sentence.\\
\\
If you cannot find the answer, respond with "I don't know".\\
\\
If the evidence is part of a paragraph, return the bounding box of the entire paragraph.\\
If the evidence is part of a table, return the bounding box of the entire table.\\
If the evidence is part of a figure, return the bounding box of the entire figure.\\
\\
User's query:\\
\{query\}
    \end{tcolorbox}
    \vspace{-10pt}
    \caption{Grounded answer generation prompt used for the InternVL family.}
    \label{fig:internvl_answer_box}
    \vspace{-0.3cm}
\end{figure*}

\begin{figure*}[t]
    \centering
    \begin{tcolorbox}[
        colback=promptbg,
        colframe=promptborder,
        arc=2mm,
        boxrule=1pt,
        width=\linewidth,
        left=5pt, right=5pt, top=5pt, bottom=5pt,
        fontupper=\ttfamily\scriptsize
    ]
     
{\sffamily\bfseries System:} Given a document image, your task is to answer the question and provide the bounding box of the answer.\\

{\sffamily\bfseries User:} Question: \{query\}
    \end{tcolorbox}
    \vspace{-10pt}
    \caption{Grounded answer generation prompt used for the VISA model.}
    \label{fig:visa_answer_box}
    \vspace{-0.3cm}
\end{figure*}

\begin{figure*}[t]
    \centering
    \begin{tcolorbox}[
        colback=promptbg,
        colframe=promptborder,
        arc=2mm,
        boxrule=1pt,
        width=\linewidth,
        left=5pt, right=5pt, top=5pt, bottom=5pt,
        fontupper=\ttfamily\scriptsize
    ]
     
{\sffamily\bfseries System:} You are an agent excellent at identifying evidence in a document page.\\
Your job is to locate the evidence used to answer the user's query on the provided image.\\
When you answer, provide only the evidence bounding box in the format <box> x1 y1 x2 y2 </box> where (x1, y1) is the top-left corner.\\
\\
If you cannot find the answer, respond with the empty box <box> </box>.\\
\\
If the evidence is part of a paragraph, return the bounding box of the entire paragraph.\\
If the evidence is part of a table, return the bounding box of the entire table.\\
If the evidence is part of a figure, return the bounding box of the entire figure.\\

{\sffamily\bfseries User:} Query:\\
\{query\}\\
\\
Answer:\\
\{answer\}
    \end{tcolorbox}
    \vspace{-10pt}
    \caption{Post-hoc grounding prompt used for the Qwen family.}
    \label{fig:qwen_box_from_asnwer}
    \vspace{-0.3cm}
\end{figure*}

\begin{figure*}[t]
    \centering
    \begin{tcolorbox}[
        colback=promptbg,
        colframe=promptborder,
        arc=2mm,
        boxrule=1pt,
        width=\linewidth,
        left=5pt, right=5pt, top=5pt, bottom=5pt,
        fontupper=\ttfamily\scriptsize
    ]
     
{\sffamily\bfseries User:} You are an agent excellent at identifying evidence in a document page.\\
Your job is to locate the evidence used to answer the user's query on the provided image.\\
When you answer, provide only the evidence bounding box in the format [x1,y1,x2,y2] where (x1, y1) is the top-left corner.\\
\\
If you cannot find the answer, respond with the empty box [].\\
\\
If the evidence is part of a paragraph, return the bounding box of the entire paragraph.\\
If the evidence is part of a table, return the bounding box of the entire table.\\
If the evidence is part of a figure, return the bounding box of the entire figure.\\
\\
User's query:\\
\{query\}\\
\\
Answer:\\
\{answer\}
    \end{tcolorbox}
    \vspace{-10pt}
    \caption{Post-hoc grounding prompt used for the InternVL family.}
    \label{fig:intervl_box_from_answer}
    \vspace{-0.3cm}
\end{figure*}

\begin{figure*}[t]
    \centering
    \begin{tcolorbox}[
        colback=promptbg,
        colframe=promptborder,
        arc=2mm,
        boxrule=1pt,
        width=\linewidth,
        left=5pt, right=5pt, top=5pt, bottom=5pt,
        fontupper=\ttfamily\scriptsize
    ]
     
{\sffamily\bfseries System:} You are an agent excellent at identifying evidence in a document page.\\
Your job is to locate the evidence used to answer the user's query on the provided image.\\
When you answer, provide only the evidence bounding box in the format <box> x1 y1 x2 y2 </box> where (x1, y1) is the top-left corner.\\
\\
If you cannot find the answer, respond with the empty box <box> </box>.\\
\\
If the evidence is part of a paragraph, return the bounding box of the entire paragraph.\\
If the evidence is part of a table, return the bounding box of the entire table.\\
If the evidence is part of a figure, return the bounding box of the entire figure.\\

{\sffamily\bfseries User:} Query:\\
\{query\}\\
\\
Answer:\\
\{answer\}
    \end{tcolorbox}
    \vspace{-10pt}
    \caption{Post-hoc grounding prompt used for the VISA model.}
    \label{fig:visa_box_from_answer}
    \vspace{-0.3cm}
\end{figure*}

\begin{figure*}[t]
    \centering
    \begin{tcolorbox}[
        colback=promptbg,      
        colframe=promptborder, 
        arc=2mm,               
        boxrule=1pt,           
        width=\linewidth,      
        left=5pt, right=5pt, top=5pt, bottom=5pt, 
         fontupper=\ttfamily\scriptsize
    ]
     
{\sffamily\bfseries System:} You are an agent excellent at identifying evidence in a document page.\\
Your job is to locate the answer to the user's query on the provided image.\\
When you answer, provide only the evidence bounding box in the format <box> x1 y1 x2 y2 </box> where (x1, y1) is the top-left corner.\\
\\
If you cannot find the answer, respond with the empty box <box> </box>.\\
\\
If the evidence is part of a paragraph, return the bounding box of the entire paragraph.\\
If the evidence is part of a table, return the bounding box of the entire table.\\
If the evidence is part of a figure, return the bounding box of the entire figure.\\

{\sffamily\bfseries User:} \{query\}
    \end{tcolorbox}
    \vspace{-10pt}
    \caption{Answer-locating prompt used for the Qwen family.}
    \label{fig:qwenvl_box}
    \vspace{-0.3cm}
\end{figure*}

\begin{figure*}[t]
    \centering
    \begin{tcolorbox}[
        colback=promptbg,      
        colframe=promptborder,
        arc=2mm,               
        boxrule=1pt,           
        width=\linewidth,      
        left=5pt, right=5pt, top=5pt, bottom=5pt, 
        fontupper=\ttfamily\scriptsize
    ]

{\sffamily\bfseries User:} You are an agent excellent at identifying evidence in a document page.\\
Your job is to locate the answer to the user's query on the provided image.\\
When you answer, provide only the evidence bounding box in the format [x1,y1,x2,y2] where (x1, y1) is the top-left corner.\\
\\
If you cannot find the answer, respond with the empty box []\\
\\
If the evidence is part of a paragraph, return the bounding box of the entire paragraph.\\
If the evidence is part of a table, return the bounding box of the entire table.\\
If the evidence is part of a figure, return the bounding box of the entire figure.\\
\\
User's query:\\
\{query\}
    \end{tcolorbox}
    \vspace{-10pt}
    \caption{Answer-locating prompt used for the InternVL family.}
    \label{fig:internvl_box}
    \vspace{-0.3cm}
\end{figure*}

\begin{figure*}[t]
    \centering
    \begin{tcolorbox}[
        colback=promptbg,      
        colframe=promptborder, 
        arc=2mm,              
        boxrule=1pt,           
        width=\linewidth,      
        left=5pt, right=5pt, top=5pt, bottom=5pt,
         fontupper=\ttfamily\scriptsize
    ]
{\sffamily\bfseries System:} You are an agent excellent at identifying evidence in a document page.\\
Your job is to locate the answer to the user's query on the provided image.\\
When you answer, provide only the evidence bounding box in the format <box> x1 y1 x2 y2 </box> where (x1, y1) is the top-left corner.\\
\\
If you cannot find the answer, respond with the empty box <box> </box>.\\
\\
If the evidence is part of a paragraph, return the bounding box of the entire paragraph.\\
If the evidence is part of a table, return the bounding box of the entire table.\\
If the evidence is part of a figure, return the bounding box of the entire figure.\\

{\sffamily\bfseries User:} Question \{query\}
    \end{tcolorbox}
    \vspace{-10pt}
    \caption{Answer-locating prompt used for the VISA model.}
    \label{fig:visa_box}
    \vspace{-0.3cm}
\end{figure*}

\begin{table*}[h!]
  \centering
  \small
  
  \setlength{\tabcolsep}{0.2em}
  \vspace{0.2cm}
  \resizebox{\textwidth}{!}{%
  \begin{tabular}{lc cc cc cc cc cc cc cc cc}
    \toprule
     & & 
    \multicolumn{2}{c}{\textbf{DocVQA}} &
    \multicolumn{2}{c}{\textbf{VisualMRC}} &
    \multicolumn{2}{c}{\textbf{VISA}} &
    \multicolumn{2}{c}{\textbf{SlideVQA}} &
    \multicolumn{2}{c}{\textbf{VisualWebB}} &
    \multicolumn{2}{c}{\textbf{LongDocURL}} &
    \multicolumn{2}{c}{\textbf{MMLongB-Doc}}  \\
    \cmidrule(lr){3-4}  \cmidrule(lr){5-6}  \cmidrule(lr){7-8}  \cmidrule(lr){9-10}  \cmidrule(lr){11-12}  \cmidrule(lr){13-14}  \cmidrule(lr){15-16}
    \textbf{Model} & & $\mathsf{P}_\text{box}$ & $\mathsf{R}_\text{box}$ & $\mathsf{P}_\text{box}$ & $\mathsf{R}_\text{box}$ & $\mathsf{P}_\text{box}$ & $\mathsf{R}_\text{box}$ & $\mathsf{P}_\text{box}$ & $\mathsf{R}_\text{box}$ & $\mathsf{P}_\text{box}$ & $\mathsf{R}_\text{box}$ & $\mathsf{P}_\text{box}$ & $\mathsf{R}_\text{box}$ & $\mathsf{P}_\text{box}$ & $\mathsf{R}_\text{box}$ & $\mathsf{\textbf{Avg}}_\text{P}$ & $\mathsf{\textbf{Avg}}_\text{R}$ \\
    \midrule
    \rowcolor{TitleColor}
    \multicolumn{18}{l}{\textit{Zero-shot MLLMs}} \\
    InternVL2.5-2B & & 0.0 & 0.0 & 1.2 & 0.8 & 0.0 & 0.0 & 0.7 & 0.7 & 0.5 & 0.4 & 0.2 & 0.1 & 0.4 & 0.4 & 0.4 & 0.4 \\
    InternVL3-2B & & 6.0 & 4.1 & 14.1 & 5.8 & 0.0 & 0.0 & 26.0 & 21.0 & 4.2 & 3.0 & 6.3 & 3.7 & 9.4 & 7.7 & 9.4 & 6.5 \\
    InternVL3.5-2B & & 28.0 & 14.8 & 3.5 & 1.4 & 2.2 & 0.2 & 0.7 & 0.6 & 28.0 & 5.9 & 1.2 & 0.9 & 2.2 & 1.1 & 9.4 & 3.6 \\
    Qwen2.5-VL-3B & & 5.0 & 3.7 & 10.9 & 12.2 & 0.1 & 0.1 & 17.3 & 17.9 & 8.0 & 6.4 & 6.6 & 5.6 & 12.7 & 10.7 & 8.7 & 8.1 \\
    Qwen3-VL-2B & & 4.1 & 20.6 & 11.7 & 31.3 & 0.3 & 0.5 & 6.3 & 29.8 & 3.1 & 16.1 & 4.2 & 25.0 & 4.7 & 27.6 & 4.9 & 21.5 \\
    \midrule
    InternVL2.5-8B & & 0.0 & 0.0 & 4.0 & 2.2 & 0.0 & 0.0 & 3.1 & 2.5 & 1.1 & 0.8 & 0.0 & 0.0 & 1.3 & 1.1 & 1.4 & 0.9 \\
    InternVL3-8B & & 6.8 & 6.1 & 9.9 & 7.0 & 0.0 & 0.0 & 19.5 & 14.3 & 2.3 & 1.7 & 4.7 & 3.3 & 7.1 & 4.4 & 7.2 & 5.2 \\
    InternVL3.5-8B & & 22.5 & 21.5 & 6.0 & 5.0 & 0.4 & 0.4 & 4.3 & 3.7 & 25.0 & 25.0 & 3.3 & 3.1 & 2.0 & 2.2 & 9.1 & 8.7 \\
    Qwen2.5-VL-7B & & 15.6 & 4.8 & 26.6 & 6.2 & 2.0 & 0.3 & 30.6 & 12.6 & 15.8 & 5.1 & 16.8 & 7.8 & 19.4 & 9.2 & 18.1 & 6.6 \\
    Qwen3-VL-8B & & 14.8 & 14.3 & 23.7 & 25.1 & 4.1 & 4.3 & 8.6 & 10.4 & 16.7 & 16.9 & 18.2 & 21.8 & 13.6 & 16.5 & 14.3 & 15.6 \\
    \midrule
    InternVL2.5-38B & & 2.3 & 0.7 & 7.2 & 6.5 & 2.2 & 0.2 & 11.1 & 9.2 & 4.2 & 2.1 & 3.1 & 3.0 & 6.6 & 5.5 & 5.3 & 3.9 \\
    InternVL3-38B & & 4.7 & 4.6 & 6.3 & 6.5 & 1.8 & 1.7 & 10.7 & 12.2 & 2.1 & 2.1 & 3.6 & 4.6 & 6.1 & 7.7 & 5.0 & 5.6 \\
    InternVL3.5-38B & & 26.7 & 26.0 & 15.0 & 14.9 & 1.4 & 1.4 & 6.5 & 6.0 & 30.3 & 29.7 & 11.4 & 11.7 & 7.5 & 7.7 & 14.1 & 13.9 \\
    Qwen2.5-VL-32B & & 9.7 & 9.8 & 18.5 & 20.8 & 1.2 & 1.5 & 9.8 & 13.8 & 27.2 & 29.7 & 8.4 & 11.7 & 12.8 & 17.6 & 12.5 & 15.0 \\
    Qwen3-VL-32B & & 41.5 & 42.1 & 55.0 & 58.5 & 29.8 & 32.3 & 29.3 & 31.9 & 31.5 & 30.9 & 44.2 & 51.0 & 42.8 & 46.0 & 39.2 & 41.8 \\
    \midrule
    \rowcolor{TitleColor}
    \multicolumn{18}{l}{\textit{Source Attribution MLLMs}} \\
    VISA-7B & & 43.2 & 43.2 & 64.4 & 64.2 & 50.5 & 50.4 & 62.7 & 62.6 & 14.6 & 14.4 & 43.0 & 43.0 & 55.9 & 55.9 & 47.7 & 47.7 \\
    \rowcolor{myhighlight} 
    \textbf{\vsa-7B} & & 65.8 & 65.8 & 74.6 & 74.4 & 58.6 & 58.5 & 66.0 & 65.8 & 30.9 & 30.5 & 59.8 & 59.8 & 60.3 & 60.3 & 59.4 & 59.3 \\
    \rowcolor{myhighlight}
    \textbf{\vsa-8B} & & \textbf{75.3} & \textbf{75.3} & \textbf{76.8} & \textbf{76.7} & 60.8 & 60.7 & \textbf{73.4} & \textbf{73.1} & 37.1 & 36.6 & \textbf{69.4} & \textbf{69.4} & \textbf{66.8} & \textbf{66.5} & \textbf{65.7} & \textbf{65.5}\\
    \rowcolor{myhighlight}
    \textbf{\vsa-I-8B} & & 69.8 & 69.8 & 76.1 & 75.9 & \textbf{61.2} & \textbf{61.2} & 65.5 & 65.2 & \textbf{40.7} & \textbf{40.1} & 68.0 & 68.0 & 66.5 & \textbf{66.5} & 64.0 & 63.8\\
    \bottomrule
  \end{tabular}
  }
  \vspace{-0.15cm}
  \caption{Precision (\textit{P}) and recall (\textit{R}) of the bounding boxes generated for the grounded answer generation task.}
\label{tab:grounded_answer_scores}
  \vspace{-0.3cm}
\end{table*}

 \begin{table*}[h!]
  \centering
  \small
  
  \setlength{\tabcolsep}{0.2em}
  \vspace{0.2cm}
  \resizebox{\textwidth}{!}{
  \begin{tabular}{lc cc cc cc cc cc cc cc cc}
    \toprule
     & & 
    \multicolumn{2}{c}{\textbf{DocVQA}} &
    \multicolumn{2}{c}{\textbf{VisualMRC}} &
    \multicolumn{2}{c}{\textbf{VISA}} &
    \multicolumn{2}{c}{\textbf{SlideVQA}} &
    \multicolumn{2}{c}{\textbf{VisualWebB}} &
    \multicolumn{2}{c}{\textbf{LongDocURL}} &
    \multicolumn{2}{c}{\textbf{MMLongB-Doc}}  \\
    \cmidrule(lr){3-4}  \cmidrule(lr){5-6}  \cmidrule(lr){7-8}  \cmidrule(lr){9-10}  \cmidrule(lr){11-12}  \cmidrule(lr){13-14}  \cmidrule(lr){15-16}
    \textbf{Model} & & $\mathsf{P}_\text{box}^\text{QA}$ & $\mathsf{R}_\text{box}^\text{QA}$ & $\mathsf{P}_\text{box}^\text{QA}$ & $\mathsf{R}_\text{box}^\text{QA}$ & $\mathsf{P}_\text{box}^\text{QA}$ & $\mathsf{R}_\text{box}^\text{QA}$ & $\mathsf{P}_\text{box}^\text{QA}$ & $\mathsf{R}_\text{box}^\text{QA}$ & $\mathsf{P}_\text{box}^\text{QA}$ & $\mathsf{R}_\text{box}^\text{QA}$ & $\mathsf{P}_\text{box}^\text{QA}$ & $\mathsf{R}_\text{box}^\text{QA}$ & $\mathsf{P}_\text{box}^\text{QA}$ & $\mathsf{R}_\text{box}^\text{QA}$ & $\mathsf{\textbf{Avg}}_\text{P}^\text{QA}$ & $\mathsf{\textbf{Avg}}_\text{R}^\text{QA}$ \\
    \midrule
    \rowcolor{TitleColor}
    \multicolumn{18}{l}{\textit{Zero-shot MLLMs}} \\
    InternVL2.5-2B & & 0.2 & 0.2 & 0.4 & 0.4 & 0.0 & 0.0 & 0.3 & 0.2 & 0.0 & 0.0 & 0.2 & 0.1 & 0.0 & 0.0 & 0.2 & 0.1 \\
    InternVL3-2B & & 15.7 & 15.6 & 13.2 & 13.1 & 0.0 & 0.0 & 21.2 & 20.9 & 6.5 & 6.4 & 5.6 & 5.5 & 12.0 & 12.0 & 10.6 & 10.5 \\
    InternVL3.5-2B & & 16.4 & 12.0 & 2.4 & 1.6 & 0.3 & 0.3 & 0.3 & 0.2 & 12.5 & 11.0 & 1.2 & 1.2 & 2.5 & 2.5 & 5.1 & 4.1 \\
    Qwen2.5-VL-3B & & 6.0 & 3.6 & 15.8 & 13.2 & 0.4 & 0.3 & 26.8 & 19.5 & 10.3 & 6.8 & 13.8 & 9.2 & 20.8 & 14.9 & 13.4 & 9.6 \\
    Qwen3-VL-2B & & 25.8 & 27.0 & 30.3 & 30.4 & 1.3 & 1.3 & 31.1 & 34.4 & 16.1 & 16.9 & 26.2 & 26.6 & 29.3 & 36.6 & 22.9 & 24.8 \\
    \midrule
    InternVL2.5-8B & & 0.6 & 0.6 & 7.7 & 7.7 & 0.1 & 0.1 & 4.2 & 4.2 & 0.9 & 0.8 & 1.3 & 1.3 & 2.9 & 2.9 & 2.5 & 2.5 \\
    InternVL3-8B & & 17.1 & 13.7 & 11.8 & 10.4 & 0.2 & 0.2 & 17.3 & 14.9 & 12.8 & 10.6 & 6.3 & 5.2 & 9.1 & 7.2 & 10.6 & 8.9 \\
    InternVL3.5-8B & & 44.8 & 44.5 & 9.6 & 9.7 & 1.7 & 1.7 & 15.0 & 15.2 & 27.8 & 27.5 & 11.3 & 11.5 & 13.8 & 14.1 & 17.7 & 17.8 \\
    Qwen2.5-VL-7B & & 19.9 & 19.7 & 26.2 & 25.5 & 1.1 & 1.1 & 31.9 & 32.1 & 17.8 & 17.4 & 20.4 & 20.4 & 27.7 & 27.5 & 20.7 & 20.5 \\
    Qwen3-VL-8B & & 27.7 & 27.2 & 41.4 & 41.1 & 6.8 & 6.7 & 18.9 & 18.6 & 24.6 & 24.1 & 35.9 & 35.2 & 32.0 & 31.5 & 26.7 & 26.3 \\
    \midrule
    InternVL2.5-38B & & 11.3 & 11.3 & 15.6 & 15.6 & 1.4 & 1.4 & 27.9 & 27.6 & 6.9 & 6.8 & 13.5 & 13.5 & 27.1 & 26.8 & 14.8 & 14.7 \\
    InternVL3-38B & & 17.8 & 17.8 & 14.7 & 14.7 & 3.2 & 3.2 & 33.3 & 33.2 & 4.3 & 4.2 & 17.5 & 17.6 & 27.2 & 27.2 & 16.8 & 16.8 \\
    InternVL3.5-38B & & 37.9 & 34.7 & 24.3 & 22.2 & 5.1 & 4.0 & 14.0 & 10.3 & 32.2 & 29.2 & 18.5 & 10.5 & 18.2 & 11.2 & 21.5 & 17.5 \\
    Qwen2.5-VL-32B & & 19.1 & 19.1 & 34.5 & 34.7 & 4.4 & 4.4 & 27.1 & 27.6 & 29.1 & 28.4 & 28.8 & 29.5 & 32.1 & 32.6 & 25.0 & 25.2 \\
    Qwen3-VL-32B & & 51.3 & 50.8 & 64.9 & 64.3 & 35.0 & 34.2 & 40.0 & 39.0 & 31.8 & 29.7 & 47.2 & 46.7 & 52.0 & 51.8 & 46.0 & 45.2 \\
    \midrule
    \rowcolor{TitleColor}
    \multicolumn{18}{l}{\textit{Source Attribution MLLMs}} \\
    VISA-7B & & 44.5 & 44.5 & 65.8 & 65.7 & 58.6 & 58.6 & 62.9 & 62.7 & 14.6 & 14.4 & 46.2 & 46.2 & 56.5 & 56.5 & 49.9 & 49.8 \\
    \rowcolor{myhighlight} 
    \textbf{\vsa-7B} & & 60.8 & 60.8 & 68.7 & 68.6 & 41.7 & 41.7 & 67.4 & 67.3 & 37.8 & 37.3 & 42.3 & 42.3 & 53.3 & 53.3 & 53.1 & 53.0\\
    \rowcolor{myhighlight}
    \textbf{\vsa-8B} & & 59.7 & 59.7 & 67.1 & 67.0 & 33.4 & 33.4 & \textbf{70.0} & \textbf{69.9} & 36.5 & 36.0 & 43.0 & 43.0 & 55.8 & 55.8 & 52.2 & 52.1 \\
    \rowcolor{myhighlight}
    \textbf{\vsa-I-8B} & & \textbf{71.5} & \textbf{71.5} & \textbf{79.3} & \textbf{79.2} & \textbf{67.1} & \textbf{67.1} & 67.8 & 67.7 & \textbf{48.1} & \textbf{47.5} & \textbf{73.0} & \textbf{73.0} & \textbf{70.3} & \textbf{70.3} & \textbf{68.2} & \textbf{68.0}
    \\
    \bottomrule
  \end{tabular}
  }
    \vspace{-0.15cm}
  \caption{Precision (\textit{P}) and recall (\textit{R}) of the bounding boxes generated for the post-hoc attribution task.}
\label{tab:post_hoc_scores}
  \vspace{-0.3cm}
\end{table*}

\begin{table*}[h!]
  \centering
  \small
  \setlength{\tabcolsep}{0.2em}
  \vspace{0.2cm}
  \resizebox{\textwidth}{!}{%
  \begin{tabular}{lc cc cc cc cc cc cc cc cc}
    \toprule
     & & 
    \multicolumn{2}{c}{\textbf{DocVQA}} &
    \multicolumn{2}{c}{\textbf{VisualMRC}} &
    \multicolumn{2}{c}{\textbf{VISA}} &
    \multicolumn{2}{c}{\textbf{SlideVQA}} &
    \multicolumn{2}{c}{\textbf{VisualWebB}} &
    \multicolumn{2}{c}{\textbf{LongDocURL}} &
    \multicolumn{2}{c}{\textbf{MMLongB-Doc}}  \\
    \cmidrule(lr){3-4}  \cmidrule(lr){5-6}  \cmidrule(lr){7-8}  \cmidrule(lr){9-10}  \cmidrule(lr){11-12}  \cmidrule(lr){13-14}  \cmidrule(lr){15-16}
    \textbf{Model} & & $\mathsf{P}_\text{box}^\text{Q}$ & $\mathsf{R}_\text{box}^\text{Q}$ & $\mathsf{P}_\text{box}^\text{Q}$ & $\mathsf{R}_\text{box}^\text{Q}$ & $\mathsf{P}_\text{box}^\text{Q}$ & $\mathsf{R}_\text{box}^\text{Q}$ & $\mathsf{P}_\text{box}^\text{Q}$ & $\mathsf{R}_\text{box}^\text{Q}$ & $\mathsf{P}_\text{box}^\text{Q}$ & $\mathsf{R}_\text{box}^\text{Q}$ & $\mathsf{P}_\text{box}^\text{Q}$ & $\mathsf{R}_\text{box}^\text{Q}$ & $\mathsf{P}_\text{box}^\text{Q}$ & $\mathsf{R}_\text{box}^\text{Q}$ & $\mathsf{\textbf{Avg}}_\text{P}^\text{Q}$ & $\mathsf{\textbf{Avg}}_\text{R}^\text{Q}$ \\
    \midrule
    \rowcolor{TitleColor}
    \multicolumn{18}{l}{\textit{Zero-shot MLLMs}} \\
    InternVL2.5-2B & & 0.0 & 0.0 & 1.5 & 0.9 & 0.2 & 0.2 & 0.7 & 0.5 & 0.0 & 0.0 & 0.3 & 0.1 & 0.6 & 0.4 & 0.5 & 0.3 \\
    InternVL3-2B & & 8.1 & 7.9 & 12.4 & 12.1 & 0.0 & 0.0 & 14.0 & 13.7 & 4.3 & 4.2 & 3.5 & 3.3 & 6.6 & 6.5 & 7.0 & 6.8 \\
    InternVL3.5-2B & & 13.6 & 9.4 & 1.0 & 0.7 & 0.5 & 0.2 & 0.2 & 0.2 & 6.3 & 3.8 & 0.2 & 0.1 & 1.2 & 1.1 & 3.3 & 2.2 \\
    Qwen2.5-VL-3B & & 10.3 & 4.5 & 16.1 & 13.6 & 0.1 & 0.1 & 33.8 & 20.3 & 10.0 & 6.8 & 12.0 & 6.2 & 18.9 & 10.5 & 14.5 & 8.9 \\
    Qwen3-VL-2B & & 28.4 & 28.1 & 27.7 & 27.7 & 1.3 & 1.3 & 28.9 & 33.3 & 16.0 & 15.7 & 27.5 & 28.5 & 26.0 & 34.4 & 22.3 & 24.1 \\
    \midrule
    InternVL2.5-8B & & 0.2 & 0.2 & 7.6 & 7.4 & 0.3 & 0.2 & 4.1 & 4.0 & 0.9 & 0.8 & 0.6 & 0.6 & 1.2 & 1.1 & 2.1 & 2.0 \\
    InternVL3-8B & & 16.1 & 13.3 & 11.8 & 8.6 & 1.0 & 0.1 & 28.4 & 19.6 & 10.9 & 6.8 & 9.5 & 7.0 & 12.6 & 8.7 & 12.9 & 9.1 \\
    InternVL3.5-8B & & 29.8 & 28.7 & 10.2 & 9.9 & 2.0 & 1.9 & 17.6 & 17.5 & 25.2 & 24.6 & 9.8 & 9.7 & 12.3 & 11.6 & 15.3 & 14.8 \\
    Qwen2.5-VL-7B & & 17.0 & 16.7 & 22.5 & 21.6 & 1.2 & 1.1 & 25.9 & 25.7 & 13.3 & 12.7 & 16.6 & 17.0 & 20.2 & 19.9 & 16.7 & 16.4 \\
    Qwen3-VL-8B 8.6 & & 17.8 & 17.3 & 31.8 & 31.1 & 5.8 & 5.7 & 12.6 & 12.2 & 16.0 & 15.7 & 27.4 & 25.6 & 23.1 & 22.5 & 19.2 & 18.6 \\
    \midrule
    InternVL2.5-38B & & 10.7 & 10.7 & 16.4 & 16.3 & 1.2 & 1.1 & 27.8 & 27.6 & 6.4 & 6.4 & 12.8 & 12.8 & 29.7 & 29.4 & 15.0 & 14.9 \\
    InternVL3-38B & & 19.4 & 19.3 & 16.5 & 16.4 & 3.1 & 3.1 & 38.0 & 37.5 & 3.9 & 3.8 & 17.0 & 16.9 & 26.7 & 26.4 & 17.8 & 17.6 \\
    InternVL3.5-38B & & 28.9 & 28.3 & 25.7 & 24.5 & 4.4 & 4.1 & 11.2 & 9.7 & 28.4 & 27.5 & 15.3 & 12.9 & 13.2 & 11.2 & 18.1 & 16.9 \\
    Qwen2.5-VL-32B & & 15.0 & 14.6 & 27.7 & 27.6 & 3.9 & 3.9 & 20.4 & 20.3 & 27.2 & 26.7 & 18.1 & 17.7 & 22.8 & 22.1 & 19.3 & 19.0 \\
    Qwen3-VL-32B & & 47.8 & 47.1 & 61.4 & 60.7 & 26.2 & 25.6 & 35.8 & 34.4 & 28.9 & 27.5 & 44.7 & 43.8 & 55.0 & 54.4 & 42.8 & 41.9 \\
    \midrule
    \rowcolor{TitleColor}
    \multicolumn{18}{l}{\textit{Source Attribution MLLMs}} \\
    VISA-7B & & 41.3 & 41.3 & 65.5 & 65.3 & 49.8 & 49.8 & 62.8 & 62.6 & 13.7 & 13.6 & 42.7 & 42.7 & 55.1 & 55.1 & 47.3 & 47.2 \\
    \rowcolor{myhighlight} 
    \textbf{\vsa-7B} & & 57.6 & 57.6 & 65.0 & 64.8 & 36.2 & 36.2 & 65.4 & 65.2 & 29.6 & 29.2 & 41.3 & 41.3 & 50.0 & 50.0 & 49.3 & 49.2\\
    \rowcolor{myhighlight}
    \textbf{\vsa-8B} & & 61.7 & 61.7 & 68.0 & 67.9 & 39.3 & 39.3 & \textbf{69.7} & \textbf{69.5} & 34.3 & 33.9 & 46.2 & 46.2 & 56.5 & 56.5 & 53.7 & 53.6 \\
    \rowcolor{myhighlight}
    \textbf{\vsa-I-8B} & & \textbf{68.7} & \textbf{68.7} & \textbf{75.6} & \textbf{75.5} & \textbf{57.9} & \textbf{57.9} & 64.3 & 64.1 & \textbf{41.6} & \textbf{41.1} & \textbf{56.1} & \textbf{56.1} & \textbf{67.8} & \textbf{67.8} & \textbf{61.7} & \textbf{61.6} \\
    \bottomrule
  \end{tabular}
  }
    \vspace{-0.15cm}
  \caption{Precision (\textit{P}) and recall (\textit{R}) of the bounding boxes generated for the answer-locating task.}
\label{tab:answer_locating_scores}
  \vspace{-0.3cm}
\end{table*}

\newpage
\bibliography{egbib}